\documentclass{article} 
\usepackage{iclr2026_conference,times}

\usepackage[utf8]{inputenc} 
\usepackage[T1]{fontenc}    
\usepackage[backref=page]{hyperref}       
\usepackage{url}            
\usepackage{booktabs}       
\usepackage{amsfonts}       
\usepackage{nicefrac}       
\usepackage{microtype}      
\usepackage[table, dvipsnames]{xcolor}  
\usepackage{graphicx}
\usepackage[inline]{enumitem}
\usepackage{subcaption}
\usepackage{amsmath}
\usepackage[capitalize,sort&compress]{cleveref}
\usepackage{multirow}
\usepackage{multicol}
\usepackage{siunitx}

\hypersetup{
    colorlinks,
    linkcolor={Red},
    citecolor={NavyBlue},
    urlcolor={Magenta}
}

\creflabelformat{equation}{#2\textup{#1}#3}
\Crefname{figure}{Fig.}{Figs.}
\Crefname{table}{Tab.}{Tabs.}
\Crefname{section}{Sec.}{Secs.}
\Crefname{appendix}{App.}{Apps.}
\Crefname{equation}{Eq.}{Eqs.}

\usepackage{pifont}
\newcommand{\cmark}{\ding{51}}%
\newcommand{\xmark}{\ding{55}}%

\newcommand{\nerf}{NeRF}
\newcommand{\nftovec}{\texttt{nf2vec}}
\newcommand{\cardace}{Cardace et al.}
\newcommand{\derek}{Lim et al.}
\newcommand{\pier}{Zama Ramirez et al.}
\newcommand{\ama}{Amaduzzi et al.}
\newcommand{\siglip}{SigLIP}
\newcommand{\mlp}{\texttt{MLP}}
\newcommand{\tri}{\texttt{TRI}}
\newcommand{\hash}{\texttt{HASH}}

\newcommand{\all}{\texttt{ALL}}
\newcommand{\mlpl}{\texttt{MLP-2L}}
\newcommand{\mlph}{\texttt{MLP-32H}}
\newcommand{\tril}{\texttt{TRI-2L}}
\newcommand{\trih}{\texttt{TRI-32H}}
\newcommand{\triw}{\texttt{TRI-16W}}
\newcommand{\tric}{\texttt{TRI-8C}}
\newcommand{\hashl}{\texttt{HASH-2L}}
\newcommand{\hashh}{\texttt{HASH-32H}}
\newcommand{\hashn}{\texttt{HASH-3N}}
\newcommand{\hasht}{\texttt{HASH-11T}}
\newcommand{\mlpob}{$\texttt{MLP}^\texttt{OB}$}
\newcommand{\triob}{$\texttt{TRI}^\texttt{OB}$}
\newcommand{\hashob}{$\texttt{HASH}^\texttt{OB}$}
\newcommand{\lr}{$\mathcal{L}_\text{R}$}
\newcommand{\lc}{$\mathcal{L}_\text{C}$}
\newcommand{\lrc}{$\mathcal{L}_{\text{R}+\text{C}}$}
\newcommand{\shapenet}{ShapenetRender}
\newcommand{\llana}{LLaNA}

\title{Weight Space Representation Learning\\on Diverse NeRF Architectures}

\author{Francesco Ballerini, Pierluigi Zama Ramirez, Luigi Di Stefano, Samuele Salti \\
University of Bologna, Italy \\
\texttt{francesco.ballerini4@unibo.it}}

\iclrfinalcopy 
\begin{document}

\maketitle

\begin{abstract}
  Neural Radiance Fields (NeRFs) have emerged as a groundbreaking paradigm for representing 3D objects and scenes by encoding shape and appearance information into the weights of a neural network. Recent studies have demonstrated that these weights can be used as input for frameworks designed to address deep learning tasks; however, such frameworks require NeRFs to adhere to a specific, predefined architecture. In this paper, we introduce the first framework capable of processing NeRFs with diverse architectures and performing inference on architectures unseen at training time. We achieve this by training a Graph Meta-Network within an unsupervised representation learning framework, and show that a contrastive objective is conducive to obtaining an architecture-agnostic latent space. In experiments conducted across 13 NeRF architectures belonging to three families (MLPs, tri-planes, and, for the first time, hash tables), our approach demonstrates robust performance in classification, retrieval, and language tasks involving multiple architectures, even unseen at training time, while also matching or exceeding the results of existing frameworks limited to single architectures. Our code and data are available at \href{https://cvlab-unibo.github.io/gmnerf}{this https URL}.
\end{abstract}

\section{Introduction}
\label{sec:intro}

Neural Radiance Fields (\nerf{}s) \citep{mildenhall2020nerf} have emerged over the last few years as a new paradigm for representing 3D objects and scenes \citep{xie2022neural}. A \nerf{} is a neural network trained on a collection of images to map 3D coordinates to color and density values, which can then be used to synthesize novel views of the underlying object or scene via volume rendering.
Due to their continuous nature, \nerf{}s can encode an arbitrary number of images at any resolution into a finite number of neural network weights, thus decoupling the number of observations and their spatial resolution from the memory required to store the 3D representation. As a result, \nerf{}s hold the potential to become a standard tool for storing and communicating 3D information, as supported by the recent publication of several \nerf{} datasets \citep{de2023scannerf, hu2023nerf, ramirez2024deep, cardace2024neural, amaduzzi2025scaling}.

With the rise of \nerf{}s as a new data format, whether and how it is possible to perform traditional deep learning tasks on them has become an increasingly relevant research question. The naive solution to this problem involves rendering views of the underlying object from its \nerf{} representation and leveraging existing neural architectures designed to process images. However, this procedure requires additional computation time and several decisions that are likely to impact its outcome, such as the number of views to render, their viewpoint, and their resolution. A more elegant and efficient approach, explored in recent works, relies on performing tasks on \nerf{}s by processing their weights as input, thereby requiring no rendering step.
This strategy is adopted by \nftovec{} \citep{ramirez2024deep}, a representation learning framework that learns to map \nerf{} weights to latent vectors by minimizing a rendering loss, where \nerf{}s are standard Multi-Layer Perceptrons (MLPs). These vectors are then used as input to deep learning pipelines for downstream tasks.
A related approach is proposed by \citet{cardace2024neural}, who, instead of employing traditional MLPs, leverage tri-planar \nerf{}s \citep{chan2022efficient}, where input coordinates are projected onto three orthogonal planes of learnable features to compute the input for an MLP.

Both \nftovec{} \citep{ramirez2024deep} and the method by \citet{cardace2024neural}, however, are designed to ingest a specific type of \nerf{} architecture (i.e.\ MLPs with fixed hidden dimensions in \nftovec{} and tri-planes with fixed spatial resolution in \cardace{}), and are thus unable to process diverse input architectures. In the context of a research domain where new \nerf{} designs are constantly being explored \citep{xie2022neural}, this assumption strongly limits their applicability. The issue of handling arbitrary architectures has recently been studied in the broader research field on \emph{meta-networks}, i.e.\ neural networks that process other neural networks as input. Specifically, Graph Meta-Networks (GMNs) have been proposed \citep{lim2024graph, kofinas2024graph}, namely Graph Neural Networks (GNNs) that can ingest any neural architecture as long as it can be first converted into a graph. Yet, these works do not experiment with \nerf{}s as input to GMNs.

\begin{figure}
    \centering
    \includegraphics[width=\linewidth]{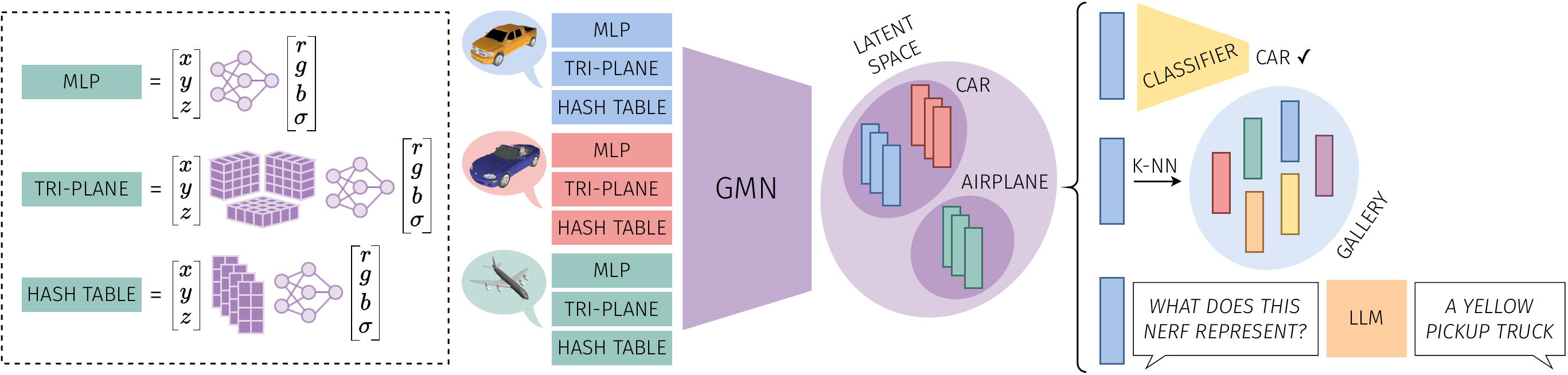}
    \caption{\textbf{Framework overview.} Our representation learning framework leverages a Graph Meta-Network \citep{lim2024graph} encoder to map weights of \nerf{}s with diverse architectures to a latent space where \nerf{}s representing similar objects are close to each other, regardless of their architecture. The embeddings are then used as input to downstream pipelines for classification, retrieval, and language tasks.}
    \label{fig:teaser}
\end{figure}

Motivated by the potential of GMNs to process diverse \nerf{} architectures, we investigate whether a GMN encoder can learn a latent space 
where distances reflect the similarity between the actual content of the radiance fields rather than their specific neural parameterization. Our empirical study reveals that this latent space organization cannot be achieved by solely relying on a rendering loss \citep{ramirez2024deep}, as such a loss alone causes different \nerf{} architectures to aggregate into distinct clusters in the embedding space, even when they represent the same underlying object. To overcome this limitation, we draw inspiration from the contrastive learning literature and introduce a \siglip{} loss term \citep{zhai2023sigmoid} that places pairs of \nerf{}s with different architectures representing the same object close to each other in latent space while pushing other pairs further apart. Combined with a rendering loss, this approach enforces an architecture-agnostic embedding space organized by class and instance.
Our experimental study features three families of popular \nerf{} architectures, namely MLPs, tri-planes \citep{chan2022efficient}, and multi-resolution hash tables \citep{muller2022instant}, for a total of 13 diverse architectures, and reveals that our encoder produces latent representations that serve as effective inputs for downstream tasks such as classification, retrieval, and language tasks.  
Notably, our framework is the first to perform tasks on \nerf{} parameterized as hash tables by processing their weights.
\cref{fig:teaser} outlines the key components of our framework.

Our contributions can be summarized as follows:
{\setlist{nosep}
\begin{itemize}[leftmargin=*]
    \item We present the first framework that performs tasks on NeRFs parameterized by diverse architectures by processing their weights.
    \item We propose to use a contrastive learning objective to create a latent space where \nerf{}s with similar content are close to each other, regardless of their architecture.
    \item We tackle, for the first time, downstream tasks on \nerf{}s parameterized by hash tables.
    \item We show that, within the families seen at training time, our framework can effectively process unseen \nerf{} architectures and generalize to \nerf{}s trained on unseen data.
    \item We achieve comparable or superior results to previous methods operating on single architectures.
\end{itemize}}

\section{Related Work}
\label{sec:related}

\textbf{Neural Radiance Fields.}
\nerf{}s were first introduced by \citet{mildenhall2020nerf}\ as a method for novel view synthesis, namely the task of generating previously unseen views of a scene from a set of sparse input images taken from different viewpoints. In the original formulation, a \nerf{} is an MLP that parameterizes a function $(x,y,z,\theta,\phi)\mapsto(r,g,b,\sigma)$ that maps a 3D position $(x,y,z)$ and a 2D viewing direction $(\theta,\phi)$ to an emitted color $(r,g,b)$ and volume density $\sigma$.
Since then, several architectural variants have been proposed, many of which combine the MLP with a trainable discrete data structure that quantizes the space of input coordinates and maps them to a higher-dimensional vector, which then serves as the actual MLP input. These structures include voxel grids \citep{liu2020neural}, tri-planes \citep{chan2022efficient}, and multi-resolution hash tables \citep{muller2022instant}, and typically result in \nerf{} architectures that can be trained much faster to convergence without sacrificing rendering quality. In particular, hash tables are the most widely adopted \nerf{} architecture in recent works \citep{barron2023zip, wang2024hyb, hu2024ngp, lee2025mfnerf, chen2024far}.
This paper focuses on the \nerf{} architectures used in \citet{ramirez2024deep} and \citet{cardace2024neural}, consisting of a single MLP and a tri-plane followed by an MLP, respectively, and the one proposed by \citet{muller2022instant}, i.e.\ a multi-resolution hash table followed by an MLP. In all three cases, the MLP adheres to the simplified formulation by \citet{mildenhall2020nerf}\ with no viewing direction, i.e.\ $(x,y,z)\mapsto(r,g,b,\sigma)$.

\textbf{Meta-networks.} 
Due to the high dimensionality of the weight space, its symmetries \citep{hecht1990algebraic}, and the impact of randomness on the solution where training converges \citep{entezari2022the, ainsworth2023git}, processing neural network weights presents unique challenges that set them apart from more common input formats. The first works to address the design of neural networks that ingest the weights of other neural networks leverage group theory to devise architectures that are equivariant to the permutation symmetries of the input networks \citep{navon2023equivariant, zhou2023permutation, zhou2023neural}. Yet, these meta-networks are tailored to specific input networks, such as MLPs and CNNs without normalization layers, and cannot generalize to arbitrary input architectures. 
To overcome this limitation, Graph Meta-Networks (GMNs) were introduced \citep{lim2024graph, kofinas2024graph}. Since GMNs are graph neural networks, they are, by design, equivariant to the node permutations of input graphs and can ingest any graph. Therefore, the challenge of processing neural network weights turns into the task of transforming the input network into a graph.
In this paper, we use the GMN formulation by \citet{lim2024graph}\ and devise our own conversion of hash tables into graphs.

\textbf{Meta-networks for \nerf{} processing.}
As the meta-network literature is still in its infancy, none of the aforementioned works include \nerf{}s as input in their experimental evaluation and instead choose to focus on simpler neural networks. The first methods to perform tasks on \nerf{}s by ingesting their weights are \nftovec{} \citep{ramirez2024deep} and the framework by \citet{cardace2024neural}. \nftovec{} is an encoder-decoder architecture trained end-to-end with a rendering loss; at inference time, the encoder takes the weights of a \nerf{} as input and produces an embedding, which in turn becomes the input to traditional deep learning pipelines for downstream tasks. More recent works \citep{ballerini2024connecting, amaduzzi2024llana} investigate the potential applications of this approach to language-related tasks. While \nftovec{} is designed to ingest MLPs, \cardace{}\ processes tri-planar \nerf{}s \citep{chan2022efficient} by discarding the MLP and giving the tri-planar component alone as input to a Transformer. Yet, both \nftovec{} and \cardace{}\ suffer from the same drawback as the first meta-networks: they are designed to handle specific \nerf{} architectures. In this work, we instead present the first architecture-agnostic framework for \nerf{} processing. 

\textbf{Contrastive learning.} 
Contrastive learning is a representation learning approach that trains models to distinguish between similar (positive) and dissimilar (negative) data pairs by aiming to embed similar data points closer together while pushing dissimilar ones farther apart in latent space \citep{chen2020simple,he2020momentum}.
Multimodal vision-language models extend this concept to align image and text modalities by maximizing the similarity between matching image-text pairs and minimizing it for mismatched ones, as demonstrated in CLIP \citep{radford2021learning}, a model that learns a shared embedding space supporting zero-shot transfer to diverse vision tasks. \citet{zhai2023sigmoid}\ build upon this foundation and propose to replace the softmax-based loss used in CLIP with a simple pairwise sigmoid loss, called \siglip{}, which is shown to work better for relatively small (4k--32k) batch sizes. For this reason, in this paper, we use the \siglip{} loss to align GMN embeddings of different \nerf{} architectures representing the same object.

\section{Method}
\label{sec:method}

\begin{figure}[t]
    \centering
    \includegraphics[width=\linewidth]{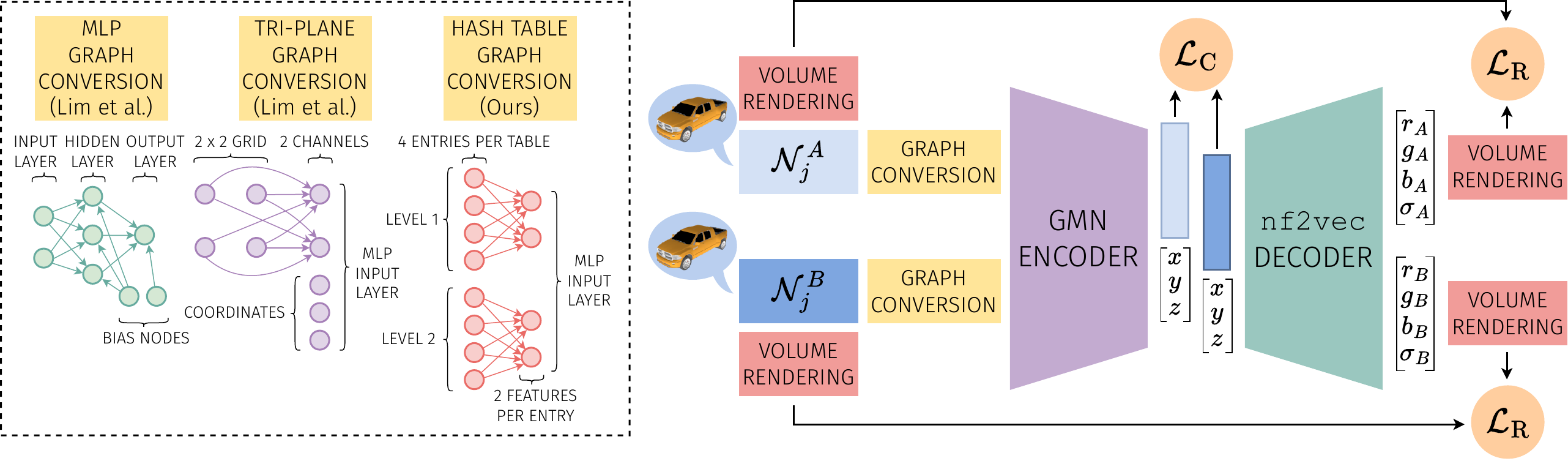}
    \caption{\textbf{Method overview.} \textbf{Left:} parameter graph construction for an MLP (left), a tri-plane (middle), and a multi-resolution hash table (right). For better clarity, the graphs of a single $2\times2\times2$ plane and of two $4\times2$ hash tables are shown. \textbf{Right:} our framework leverages a  Graph Meta-Network \citep{lim2024graph}  encoder alongside  the \nftovec{} decoder \citep{ramirez2024deep} and is trained end-to-end on a dataset of \nerf{}s with different architectures ($\mathcal{N}_j^A$, $\mathcal{N}_j^B$) with both a rendering (\lr) and a contrastive (\lc) loss.}
    \label{fig:method}
\end{figure}

We tackle the challenging problem of embedding NeRFs parameterized by different neural architectures with a representation learning framework. Our method uses an encoder and a decoder trained end-to-end with a combination of rendering and contrastive objectives, where the encoder is implemented as a Graph Meta-Network (GMN). After training, the frozen encoder converts \nerf{} weights into an embedding that can serve as input to deep learning pipelines for downstream tasks. In the remainder of this section, we will describe each framework component; further details are provided in \cref{app:details}.

\textbf{From \nerf{}s to graphs.} In order for a \nerf{} to be ingested by the encoder, it must be converted into a graph. The naive approach to perform this conversion would be to adopt the standard computation graph formulation, namely representing a neural network as a Directed Acyclic Graph (DAG), where nodes are activations and edges hold weight values. However, computation graphs scale poorly with the number of activations in networks with weight-sharing schemes, as a single weight requires multiple edges, one for each activation it affects. This limitation has motivated \citet{lim2024graph} to introduce the \emph{parameter graph} representation, where each weight is associated with a single edge of the graph, rather than multiple edges. 
\derek{}\ describe the parameter graph construction of several common neural layers, which can then be concatenated to form the overall parameter graph. Among these layers, they provide the parameter graph conversion of spatial parameter grids, such as those often used in \nerf{} architectures, specifically tri-planar ones \citep{chan2022efficient}. However, they do not extend their parameter graph formulation to multi-resolution hash tables \citep{muller2022instant}. In this paper, we leverage \derek's graph representations of linear layers and tri-planes, detailed in \cref{app:graph}, and propose a conversion into parameter graphs for hash tables, described here. These three graph representations are shown in \cref{fig:method} (left). 
Specifically, hash tables map spatial locations of a dense 3D grid to feature vectors stored in a table via a hashing function; in practice, multiple separate hash tables are used to index grids at different resolutions (hence the \emph{multi-resolution} terminology). To compute the output, the feature vectors of the grid vertices enclosing a query point $(x,y,z)$ are interpolated at each resolution, and the results are concatenated to serve as input to an MLP. The most direct way to convert a hash table into a parameter graph would be to explicitly model the 3D grids of features encoded by the tables and adopt a graph representation analogous to the one used for tri-planes. Albeit simple, explicitly modeling the underlying voxel grid would require as many nodes as there are spatial locations, i.e.\ it would scale cubically with the resolution. Instead, we construct the parameter graph of a hash table with a node for each table entry and a node for each feature vector dimension. Then, we connect each entry node to each feature node via an edge that stores the corresponding feature value. This subgraph construction is repeated for each hash table, i.e.\ for each resolution level. Finally, the concatenation of all feature nodes becomes the first layer of the MLP parameter graph that follows. This graph representation preserves the memory efficiency of the hash table and is the one we adopt.

\textbf{Encoder.} Our framework's encoder is the GMN proposed by \citet{lim2024graph}, i.e.\ a standard message-passing Graph Neural Network (GNN) \citep{battaglia2018relational} with node and edge features but no global graph-level feature. Node features are updated by message-passing along neighbors and contribute to updating the features of the edges connecting them. The final embedding is obtained via an average pooling of the edge features. Notably, as the encoder is a GNN, it can process any input graph and, hence, any neural network that has undergone the previously described parameter-graph conversion, thereby allowing our framework to handle any \nerf{} architecture for which a graph representation is known.

\textbf{Decoder.} Our framework leverages the decoder first introduced by \nftovec{} \citep{ramirez2024deep}, which takes as input the concatenation of the embedding produced by the encoder alongside a frequency encoding \citep{mildenhall2020nerf} of a 3D point $(x,y,z)$ and outputs a learned approximation of the radiance field value $(r,g,b,\sigma)$ at that point. Thus, the combination of decoder and embedding can itself be seen as a conditioned neural radiance field. Inspired by \citet{park2019deepsdf}, the decoder architecture consists of a simple succession of linear layers, intertwined with ReLU activations, and a single skip connection from the input to halfway through the network.

\textbf{Training.} The encoder and the decoder are trained end-to-end with a combination of two loss terms: a rendering loss and a contrastive loss.
The rendering loss is the one used by \citet{ramirez2024deep} to train the \nftovec{} framework and can be described as follows. Consider a \nerf{} $\mathcal{N}\colon\mathbf{x}\mapsto(\mathbf{c},\sigma)$ with $\mathbf{x}=(x,y,z)$ and $\mathbf{c}=(r,g,b)$ and let $I\in\mathcal{I}$ be one of the images $\mathcal{N}$ was trained on, with corresponding camera pose and intrinsic parameters. Let $\mathcal{X}=\{\mathbf{x}_i\}$ be the set of points sampled along a ray cast from image $I$ into the scene and passing through pixel $\mathbf{p}$ on the image plane, and let $C_\mathcal{N}(\mathbf{p})$ be the color computed via volume rendering by accumulating the contributions of the $(\mathbf{c}_i,\sigma_i)$ output values of $\mathcal{N}$ for all inputs $\mathbf{x}_i\in\mathcal{X}$. Analogously, let $C_\text{D}(\mathbf{p})$ be the color computed with the output values produced by the decoder when the encoder takes $\mathcal{N}$ as input. The rendering loss associated with \nerf{} $\mathcal{N}$ is  defined as
\begin{equation}\label{eq:rend}
    \mathcal{L}_\text{R}(\mathcal{N})=\sum_{I\in\mathcal{I}}\sum_{\mathbf{p}\in \mathcal{S}(I)}\operatorname{smoothL1}(C_\mathcal{N}(\mathbf{p}), C_\text{D}(\mathbf{p}))
\end{equation}
where $\mathcal{S}(\mathcal{I})$ is a subset of pixels sampled from $\mathcal{I}$ and $\operatorname{smoothL1}$ is the smooth L1 loss \citep{girshick2015fast}. More precisely, \pier{}\ compute one rendering loss term for foreground and one for background pixels, and express $\mathcal{L}_\text{R}$ as a weighted sum of the two.
Let $\mathcal{M},\mathcal{T},\mathcal{H}$ be three \nerf{} architectures and consider a dataset of \nerf{}s of 3D objects where each object $j$ appears three times: as a \nerf{} $\mathcal{M}_j$ parameterized by $\mathcal{M}$, as a \nerf{} $\mathcal{T}_j$ parameterized by $\mathcal{T}$, and as a \nerf{} $\mathcal{H}_j$ parameterized by $\mathcal{H}$. Given a mini-batch $\mathcal{B}=\{(\mathcal{N}_1^A,\mathcal{N}_1^B),(\mathcal{N}_2^A,\mathcal{N}_2^B),\dots\}$, where each $(\mathcal{N}_j^A,\mathcal{N}_j^B)$ is randomly sampled from $\{(\mathcal{M}_j,\mathcal{T}_j),(\mathcal{T}_j,\mathcal{H}_j),(\mathcal{M}_j,\mathcal{H}_j)\}$, the rendering loss computed over  $\mathcal{B}$ is the average rendering loss of each \nerf{} $\mathcal{N}$ in $\mathcal{B}$, i.e.\
\begin{equation} \label{eq:render}
    \mathcal{L}_\text{R}=\frac{1}{2|\mathcal{B}|}\sum_{j=1}^{|\mathcal{B}|}\sum_{\mathcal{N}\in(\mathcal{N}_j^A,\mathcal{N}_j^B)}\mathcal{L}_\text{R}(\mathcal{N})
\end{equation}
where $\mathcal{L}_\text{R}(\mathcal{N})$ is  defined in \cref{eq:rend}.
Our contrastive loss follows instead the definition by \citet{zhai2023sigmoid}, namely
\begin{equation} \label{eq:contrastive}
    \mathcal{L}_\text{C}=-\frac{1}{|\mathcal{B}|}\sum_{j=1}^{|\mathcal{B}|}\sum_{k=1}^{|\mathcal{B}|}\ln\frac{1}{1+e^{-\ell_{jk}(t\mathbf{u}_j\cdot\mathbf{v}_k+b)}}
\end{equation}
where $\ell_{jk}=1$ if $j=k$ and $-1$ otherwise, $t$ and $b$ are learnable scalar hyperparameters, and $\mathbf{u}_j$ and $\mathbf{v}_k$ are the L2-normalized encoder embeddings of $\mathcal{N}_j^A$ and $\mathcal{N}_k^B$, respectively. Finally, the combined loss used to train our framework is
\begin{equation} \label{eq:combined}
    \mathcal{L}_{\text{R}+\text{C}}=\mathcal{L}_\text{R}+\lambda\mathcal{L}_\text{C}
\end{equation}
where $\lambda$ is a fixed hyperparameter. The rationale behind the choice of this loss is discussed in \cref{sec:latent}. \cref{fig:method} (right) shows an overview of our training procedure.

\textbf{Inference.} At inference time, a single forward pass of the encoder converts the parameter graph of a \nerf{} into a latent vector, which we then use as input to deep learning pipelines for classification, retrieval, and language tasks, as outlined in \cref{fig:teaser} and detailed in \cref{sec:classification,sec:retrieval,sec:language}.

\section{Experiments}
\label{sec:experiments}

\textbf{Datasets.} Our experimental evaluation is based on three families of \nerf{} architectures: MLP-based (i.e.\ a vanilla MLP), tri-planar (i.e.\ a tri-plane \citep{chan2022efficient} followed by an MLP), and hash-based (i.e.\ a multi-resolution hash table \citep{muller2022instant} followed by an MLP). These families are exemplified by the following datasets:
\begin{enumerate*}[label=(\roman*)]
    \item one consisting of \citet{ramirez2024deep}'s MLP-based \nerf{}s only, which will be referred to as \mlp{};
    \item one consisting of \citet{cardace2024neural}'s tri-planar \nerf{}s only, which will be referred to as \tri{};
    \item one, created by us, consisting of hash-based \nerf{}s only, which will be referred to as \hash{};
    \item the union of \mlp{}, \tri{}, and \hash{}, i.e.\ a dataset where each object appears three times: once as an MLP-based \nerf{}, then as a tri-planar \nerf{}, and finally as a hash-based \nerf{}; we will refer to this dataset as \all{}.
\end{enumerate*}
To further assess the ability of our framework to perform tasks on arbitrary \nerf{} architectures, we create additional test sets of \nerf{}s featuring architectures belonging to the same families of those seen at training time but with different hyperparameters, which we will refer to as \emph{unseen architectures}. The specifics of the architectures of all our datasets are reported in \cref{tab:arch}. Overall, we consider a total of 13 architectures across the three families.
Following the protocol first introduced by \citet{ramirez2024deep}, all the aforementioned \nerf{}s are trained on \shapenet{} \citep{xu2019disn}, a dataset providing RGB images of synthetic 3D objects together with their class label.

\textbf{Models.} To assess the impact of different losses on training, we introduce a distinction between three versions of our framework, depending on the learning objective:
\begin{enumerate*}[label=(\roman*)]
    \item $\mathcal{L}_\text{R}$, where the framework has been trained with the rendering loss of \cref{eq:render} alone;
    \item $\mathcal{L}_{\text{R}+\text{C}}$, where the framework has been trained with a combination of rendering and contrastive losses as in \cref{eq:combined}, with $\lambda=\num{2e-2}$ (as it leads to similar magnitudes in the two terms);
    \item $\mathcal{L}_\text{C}$, where the framework has been trained with the contrastive loss of \cref{eq:contrastive} alone.
\end{enumerate*}

\textbf{Single vs multi-architecture.} Our method is the first neural processing framework able to handle multiple \nerf{} architectures. Thus, the natural setting to test it is when it is trained on \all{}; we will refer to this scenario as the \emph{multi-architecture setting}. Yet, our method can also be used when the input \nerf{}s all share the same architecture, e.g.\ when our framework has been trained on \mlp{}, \tri{}, or \hash{} only, by dropping the contrastive loss term. Therefore, to test its generality, we also perform experiments in such a scenario, which we will refer to as the \emph{single-architecture setting}. In this setting, our approach can be compared to previous methods \citep{ramirez2024deep,cardace2024neural} that are limited to handling specific architectures.

\begin{table}[t]
    \caption{\textbf{Dataset architectures.} Architectural hyperparameters of \nerf{} datasets featured in our experiments, including those used at inference time only (\emph{unseen}).}
    \label{tab:arch}
    \centering
    \resizebox{\linewidth}{!}{
    \begin{tabular}{lrrrrrrrrrrrrr}
        \toprule
        & \multicolumn{3}{c}{Training} & \multicolumn{10}{c}{Unseen} \\
        \cmidrule(lr){2-4} \cmidrule(lr){5-14}
        & \mlp{} & \tri{} & \hash{} & \mlpl{} & \mlph{} & \tril{} & \trih{} & \triw{} & \tric{} & \hashl{} & \hashh{} & \hashn{} & \hasht{} \\
        \midrule
        MLP Hidden Layers & 3 & 3 & 3 & 2 & 3 & 2 & 3 & 3 & 3 & 2 & 3 & 3 & 3 \\
        MLP Hidden Dim & 64 & 64 & 64 & 64 & 32 & 64 & 32 & 64 & 64 & 64 & 32 & 64 & 64 \\
        Tri-plane Resolution & -- & 32 & -- & -- & -- & 32 & 32 & 16 & 32 & -- & -- & -- & -- \\
        Tri-plane Channels & -- & 16 & -- & -- & -- & 16 & 16 & 16 & 8 & -- & -- & -- & -- \\
        Hash Table Levels & -- & -- & 4 & -- & -- & -- & -- & -- & -- & 4 & 4 & 3 & 4 \\
        Hash Table Size ($\log_2$) & -- & -- & 12 & -- & -- & -- & -- & -- & -- & 12 & 12 & 12 & 11 \\
        \bottomrule
    \end{tabular}}
\end{table}

\begin{figure*}[t]
\vspace{-5pt}
    \centering
    \includegraphics[width=\linewidth]{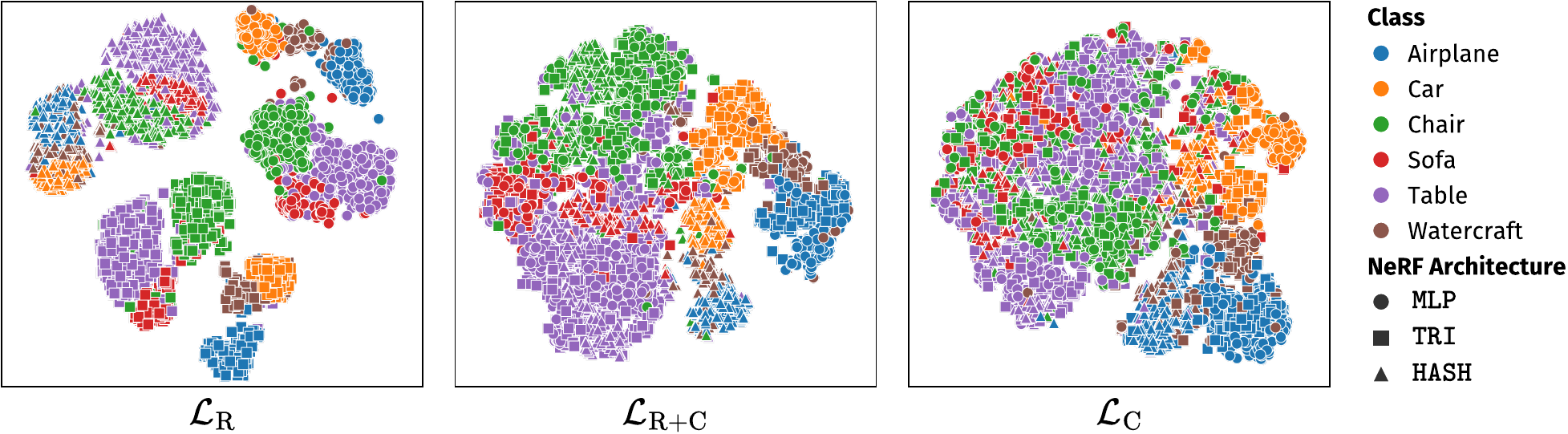}
    \caption{\textbf{t-SNE plots.} 2D projections of the latent space created by our framework when trained on a dataset of \nerf{}s of \shapenet{} objects \citep{xu2019disn}, where each object is represented by three \nerf{}s parameterized by different architectures: MLPs, tri-planes, and multi-resolution hash tables.
    \label{fig:tsne}}
\end{figure*}

\subsection{Latent Space Analysis}
\label{sec:latent}

To study the impact of different learning objectives on the organization of the resulting \nerf{} latent space, we apply t-SNE dimensionality reduction \citep{vandermaaten08vis} to the embeddings computed by our framework on \nerf{}s belonging to the test set of the \all{} dataset. \cref{fig:tsne} shows the resulting bi-dimensional plots.
Some interesting patterns can be noted. When trained to minimize \lr{} alone (\cref{fig:tsne}, left), the encoder creates an embedding space that is clustered by class, even though class labels were not used at training time. This behavior is a byproduct of the loss, which enforces \nerf{}s encoding similar shapes and colors to lie nearby in the latent space. Another outcome one may expect is that the same object represented by different \nerf{} architectures is projected into nearby embeddings; however, this turns out not to be the case. Instead, three distinct clusters emerge for each class, each corresponding to a  \nerf{} architecture, and these clusters are further away from each other than clusters sharing the same underlying architecture. This outcome shows that \lr{} alone does not directly encourage the model to align \nerf{} embeddings regardless of the input architecture.
Conversely, when trained to minimize \lc{} alone (\cref{fig:tsne}, right), the encoder significantly reduces the distance between different architectures representing the same class, but also provides a considerably less distinguishable clusterization by class compared to  \lr{} case (especially for some classes, like \emph{chairs}, \emph{sofas}, and \emph{tables}). 
Finally, \lrc{} (\cref{fig:tsne}, middle) strikes a balance between these latent space properties. Compared to \lr{}, one macro-cluster per class is present, although hash tables tend to form a separate sub-cluster (which is especially evident for \emph{airplanes} and \emph{watercrafts}). Compared to \lc{}, classes are more separated (although not as much as with \lr{}), but the sub-clusterization of hash tables is more apparent. Thus, we expect \lr{} to be the best choice for tasks where the separation between classes is highly relevant; \lrc{} and \lc{}, on the other hand, are likely to be the most effective for tasks where strong invariance to the \nerf{} architecture is required, with \lrc{} being preferable, as it preserves more separation between classes. For these reasons, the remainder of this section shows experimental results obtained with \lr{} and \lrc{}; results for \lc{} are reported in \cref{app:experiments}.

\begin{table}[t]
    \centering
    \begin{minipage}[t]{0.45\linewidth}
        \setlength{\tabcolsep}{12pt}
        \caption{\textbf{\nerf{} classification (multi-architecture).} The encoder is trained on \all{}; the classifier is trained on the datasets in column 2 and tested on those in columns 3--5.}
        \label{tab:classification-multi}
        \centering
        \resizebox{\linewidth}{!}{
        \begin{tabular}{lcrrr}
            \toprule
            && \multicolumn{3}{c}{Accuracy (\%) $\uparrow$} \\
            \cmidrule(lr){3-5}
            Method & Classifier Training Set & \mlp & \tri & \hash \\
            \midrule
            \nftovec{} & \multirow{2}{*}{--} & -- & -- & -- \\
            \citeauthor{cardace2024neural} && -- & -- & -- \\
            \cmidrule(lr){1-5}
            \lr{} (ours) & \multirow{2}{*}{\all} & \cellcolor{red!15} \textbf{93.6} & \cellcolor{red!15} \textbf{94.0} & \cellcolor{red!15} \textbf{92.4} \\
            \lrc{} (ours) && \cellcolor{red!15} 90.7 & \cellcolor{red!15} 90.6 & \cellcolor{red!15} 90.0 \\
            \cmidrule(lr){1-5}
            \lr{} (ours) & \multirow{2}{*}{\mlp} & \cellcolor{red!15} \textbf{93.8} & 25.3 & 19.3 \\
            \lrc{} (ours) && \cellcolor{red!15} 91.5 & \textbf{58.6} & \textbf{56.6} \\
            \cmidrule(lr){1-5}
            \lr{} (ours) & \multirow{2}{*}{\tri} & 11.4 & \cellcolor{red!15} \textbf{93.8} & 9.3 \\
            \lrc{} (ours) && \textbf{77.8} & \cellcolor{red!15} 91.0 & \textbf{66.5} \\
            \cmidrule(lr){1-5}
            \lr{} (ours) & \multirow{2}{*}{\hash} & 13.8 & \textbf{35.7} & \cellcolor{red!15} \textbf{92.7} \\
            \lrc{} (ours) && \textbf{54.1} & 35.5 & \cellcolor{red!15} 90.6 \\
            \bottomrule
        \end{tabular}}
    \end{minipage}
    \hspace{20pt}
    \begin{minipage}[t]{0.4\linewidth}
        \setlength{\tabcolsep}{13pt}
        \caption{\textbf{\nerf{} classification (single-architecture).} The classifier is trained and tested on the same dataset on which the encoder is trained.}
        \label{tab:classification-single}
        \centering
        \resizebox{\linewidth}{!}{
        \begin{tabular}{lcc}
            \toprule
            \multirow{2}{*}{Method} & Encoder & \multirow{2}{*}{Accuracy (\%) $\uparrow$} \\
            & Training Set & \\
            \midrule
            \nftovec{} & \multirow{3}{*}{\mlp} & 92.1 \\
            \citeauthor{cardace2024neural} && -- \\
            \lr{} (ours) && \textbf{93.6} \\
            \cmidrule(lr){1-3}
            \nftovec{} & \multirow{3}{*}{\tri} & -- \\
            \citeauthor{cardace2024neural} && 93.1 \\
            \lr{} (ours) && \textbf{94.0} \\
            \cmidrule(lr){1-3}
            \nftovec{} & \multirow{3}{*}{\hash} & -- \\
            \citeauthor{cardace2024neural} && -- \\
            \lr{} (ours) && \textbf{92.5} \\
            \bottomrule
        \end{tabular}}
    \end{minipage}
\end{table}

\subsection{\nerf{} Classification}
\label{sec:classification}

Once our framework has been trained, \nerf{} classification is performed by learning a downstream classifier $\mathcal{C}$ to predict the labels of the embeddings produced by the encoder; an overview of this procedure is shown in \cref{fig:teaser}. We report results for two cases: classification of \nerf{} architectures on which our framework was trained and classification of unseen architectures. In both cases, since the multi-architecture setting requires evaluating methods trained on \all{}, neither one of the previous works, i.e.\ \nftovec{} \citep{ramirez2024deep} and \citet{cardace2024neural}, can be applied to this scenario, as they can only be trained on \mlp{} and \tri{}, respectively.

\textbf{Training architectures.} \cref{tab:classification-multi} shows \nerf{} classification results in the multi-architecture setting on architectures used to train our framework. 
When $\mathcal{C}$ has been trained on \all{}, \lr{} performs the best: this result is consistent with the better separation between clusters corresponding to different classes provided by \lr{}, as shown in \cref{fig:tsne} (left) and discussed in \cref{sec:latent}. This trend can also be observed, as expected, when $\mathcal{C}$ is trained on either \mlp{}, \tri{}, or \hash{} and then tested on the same dataset, replicating the previous scenario with smaller datasets.
On the other hand, the introduction of a contrastive objective in \lrc{} is key to performance whenever $\mathcal{C}$ is trained on a single-architecture dataset (e.g.\ \mlp{}) and tested on a dataset with a different architecture (e.g.\ \tri{} or \hash{}), as clusters of \nerf{}s belonging to the same class but parameterized by different architectures are closer in the embedding space than with \lr{} alone. Overall, our method consistently achieves remarkable accuracies when tested on architectures included in the training set of $\mathcal{C}$ (i.e.\ red cells in \cref{tab:classification-multi}).
In the single-architecture setting, instead, previous methods can be trained and evaluated on the datasets corresponding to the \nerf{} architecture they were designed to process, whereas \lrc{} cannot be applied, as there is a single architecture and, therefore, no positive pairs to compute the contrastive loss. As shown in \cref{tab:classification-single}, when both the encoder and the classifier are trained and tested on either \mlp{} or \tri{}, our method outperforms both \nftovec{} and \cardace{}, while also achieving high accuracy on \hash{}. Hence, beyond its original aim, our approach can also provide a competitive alternative to single-architecture frameworks.

\begin{table}[t]
    \caption{\textbf{\nerf{} classification of unseen architectures (multi-architecture).} The encoder is trained on \all{}; the classifier is trained on the datasets in column 2 and tested on those in columns 3--12, containing \nerf{} architectures unseen at training time.}
    \label{tab:classification-multi-unseen}
    \centering
    \setlength{\tabcolsep}{10pt}
    \resizebox{\linewidth}{!}{
    \begin{tabular}{lcrrrrrrrrrr}
        \toprule
        && \multicolumn{10}{c}{Accuracy (\%) $\uparrow$} \\
        \cmidrule(lr){3-12}
         Method & Classifier Training Set & \mlpl & \mlph & \tril & \trih & \triw & \tric & \hashl & \hashh & \hashn & \hasht \\
        \midrule
        \nftovec{} & \multirow{2}{*}{--} & -- & -- & -- & -- & -- & -- & -- & -- & -- & -- \\
        \citeauthor{cardace2024neural} && -- & -- & -- & -- & -- & -- & -- & -- & -- & -- \\
        \cmidrule(lr){1-12}
        \lr{} (ours) & \multirow{2}{*}{\all} & \cellcolor{orange!15} \textbf{91.3} & \cellcolor{orange!15} \textbf{87.4} & \cellcolor{orange!15} \textbf{93.2} & \cellcolor{orange!15} \textbf{88.3} & \cellcolor{orange!15} 24.8 & \cellcolor{orange!15} \textbf{69.4} & \cellcolor{orange!15} \textbf{91.9} & \cellcolor{orange!15} \textbf{91.2} & \cellcolor{orange!15} \textbf{88.3} & \cellcolor{orange!15} 24.5 \\
        \lrc{} (ours) && \cellcolor{orange!15} 85.9 & \cellcolor{orange!15} 83.8 & \cellcolor{orange!15} 87.0 & \cellcolor{orange!15} 84.1 & \cellcolor{orange!15} \textbf{72.1} & \cellcolor{orange!15} 30.8 & \cellcolor{orange!15} 89.2 & \cellcolor{orange!15} 87.4 & \cellcolor{orange!15} 86.8 & \cellcolor{orange!15} \textbf{27.8} \\
        \cmidrule(lr){1-12}
        \lr{} (ours) & \multirow{2}{*}{\mlp} & \cellcolor{orange!15} \textbf{91.3} & \cellcolor{orange!15} \textbf{86.1} & 22.6 & 23.2 & 7.9 & \textbf{21.0} & 19.7 & 20.6 & 21.8 & 7.6 \\
        \lrc{} (ours) && \cellcolor{orange!15} 86.6 & \cellcolor{orange!15} 81.3 & \textbf{63.3} & \textbf{43.4} & \textbf{59.2} & 13.1 & \textbf{51.5} & \textbf{50.8} & \textbf{55.0} & \textbf{24.8} \\
        \cmidrule(lr){1-12}
        \lr{} (ours) & \multirow{2}{*}{\tri} & 11.9 & 10.6 & \cellcolor{orange!15} \textbf{92.1} & \cellcolor{orange!15} \textbf{84.8} & \cellcolor{orange!15} 32.1 & \cellcolor{orange!15} \textbf{43.6} & 9.2 & 8.8 & 15.0 & 5.2 \\
        \lrc{} (ours) && \textbf{60.7} & \textbf{58.0} & \cellcolor{orange!15} 86.9 & \cellcolor{orange!15} 83.2 & \cellcolor{orange!15} \textbf{63.9} & \cellcolor{orange!15} 30.2 & \textbf{59.8} & \textbf{61.5} & \textbf{66.3} & \textbf{28.5} \\
        \cmidrule(lr){1-12}
        \lr{} (ours) & \multirow{2}{*}{\hash} & 10.7 & 7.4 & 33.9 & \textbf{36.3} & 19.1 & \textbf{23.0} & \cellcolor{orange!15} \textbf{91.6} & \cellcolor{orange!15} \textbf{91.4} & \cellcolor{orange!15} \textbf{87.8} & \cellcolor{orange!15} \textbf{29.2} \\
        \lrc{} (ours) && \textbf{47.0} & \textbf{40.4} & \textbf{40.6} & 34.0 & \textbf{43.4} & 19.3 & \cellcolor{orange!15} 89.5 & \cellcolor{orange!15} 87.9 & \cellcolor{orange!15} 86.4 & \cellcolor{orange!15} 25.7 \\
        \bottomrule
    \end{tabular}}
\end{table}

\begin{table}[t]
    \caption{\textbf{\nerf{} classification of unseen architectures (single-architecture).} The classifier is trained on the same dataset as the encoder and tested on the datasets in columns 3--12, containing \nerf{} architectures unseen at training time.}
    \label{tab:classification-single-unseen}
    \centering
    \setlength{\tabcolsep}{10pt}
    \resizebox{\linewidth}{!}{
    \begin{tabular}{lcrrrrrrrrrr}
        \toprule
        && \multicolumn{10}{c}{Accuracy (\%) $\uparrow$} \\
        \cmidrule(lr){3-12}
         Method & Encoder Training Set & \mlpl & \mlph & \tril & \trih & \triw & \tric & \hashl & \hashh & \hashn & \hasht \\
        \midrule
        \nftovec{} & \multirow{3}{*}{\mlp} & \cellcolor{orange!15} 63.7 & \cellcolor{orange!15} -- & -- & -- & -- & -- & -- & -- & -- & -- \\
        \citeauthor{cardace2024neural} && \cellcolor{orange!15} -- & \cellcolor{orange!15} -- & -- & -- & -- & -- & -- & -- & -- & -- \\
        \lr{} (ours) && \cellcolor{orange!15} \textbf{91.8} & \cellcolor{orange!15} \textbf{83.7} & \textbf{10.9} & \textbf{6.1} & \textbf{7.0} & \textbf{4.9} & \textbf{5.7} & \textbf{5.6} & \textbf{4.0} & \textbf{4.9} \\
        \cmidrule(lr){1-12}
        \nftovec{} & \multirow{3}{*}{\tri} & -- & -- & \cellcolor{orange!15} -- & \cellcolor{orange!15} -- & \cellcolor{orange!15} -- & \cellcolor{orange!15} -- & -- & -- & -- & -- \\
        \citeauthor{cardace2024neural} && -- & -- & \cellcolor{orange!15} \textbf{93.5} & \cellcolor{orange!15} \textbf{92.0} & \cellcolor{orange!15} -- & \cellcolor{orange!15} 68.6 & -- & -- & -- & -- \\
        \lr{} (ours) && \textbf{10.1} & \textbf{10.0} & \cellcolor{orange!15} 92.6 & \cellcolor{orange!15} 82.5 & \cellcolor{orange!15} \textbf{22.9} & \cellcolor{orange!15} \textbf{72.8} & \textbf{17.4} & \textbf{13.6} & \textbf{10.8} & \textbf{12.1} \\
        \cmidrule(lr){1-12}
        \nftovec{} & \multirow{3}{*}{\hash} & -- & -- & -- & -- & -- & -- & \cellcolor{orange!15} -- & \cellcolor{orange!15} -- & \cellcolor{orange!15} -- & \cellcolor{orange!15} -- \\
        \citeauthor{cardace2024neural} && -- & -- & -- & -- & -- & -- & \cellcolor{orange!15} -- & \cellcolor{orange!15} -- & \cellcolor{orange!15} -- & \cellcolor{orange!15} -- \\
        \lr{} (ours) && \textbf{8.3} & \textbf{8.9} & \textbf{15.4} & \textbf{21.3} & \textbf{21.3} & \textbf{21.3} & \cellcolor{orange!15} \textbf{90.6} & \cellcolor{orange!15} \textbf{91.0} & \cellcolor{orange!15} \textbf{88.2} & \cellcolor{orange!15} \textbf{28.8} \\
        \bottomrule
    \end{tabular}}
\end{table}

\textbf{Unseen architectures.} Multi-architecture classification results on unseen architectures are shown in \cref{tab:classification-multi-unseen}. Remarkably, a trend coherent with that of \cref{tab:classification-multi} can be noticed, namely that \lr{} tends to perform best when the classifier $\mathcal{C}$ is trained either on \all{} or on an architecture belonging to the family of the unseen one presented at test time (i.e.\ orange cells in \cref{tab:classification-multi-unseen}), whereas \lrc{} almost always prevails when training and test families differ, usually by a significant margin. Notable exceptions are \triw{}, where \lrc{} consistently yields the best results, and \tric{}, where \lr{} outperforms \lrc{} (although less dramatically) even when changing the architecture family. Some other interesting patterns can be noted: in MLP and tri-plane families, reducing the number of hidden layers of an MLP has less impact on accuracy than reducing the hidden dimension, whereas it does not seem to have a significant effect on hash table families; furthermore, changing hash table size is the most disruptive change, even when encoder and classifier have seen hash-based \nerf{}s at training time. These patterns can serve as the basis for future investigations into this topic. 
Overall, it is worth highlighting that, in the most relevant scenario, i.e.\ when our framework has been trained on \all{}, our method is able to reach remarkable accuracies on most of the unseen variations of known architecture families it is tested on.
In the single-architecture setting, shown in \cref{tab:classification-single-unseen}, \lr{} performs much better than \nftovec{} in the only case in which the latter can be tested. Conversely, \lr{} performs worse than \citet{cardace2024neural} in \tril{} and \trih{}, i.e.\ \tri{} variations that change the MLP architecture. This is not surprising since, when performing classification, \cardace{}\ processes the tri-plane alone and discards the MLP, which makes their approach invariant to changes in the MLP architecture. Yet, since their method is tied to the tri-plane quantization, it cannot process \triw{} and is less robust than ours to the reduction in the number of channels. Finally, our method is the only one that can be tested on the hash table family, where it exhibits robust performance when trained on \hash{}, except for \hasht{}, which already emerged as particularly challenging in the results of \cref{tab:classification-multi-unseen}.

\begin{table}[t]
    \centering
    \begin{minipage}[t]{0.47\linewidth}
        \setlength{\tabcolsep}{6.5pt}
        \caption{\textbf{\nerf{} retrieval (\shapenet{}).} The encoder is trained on \all. Query/gallery combinations belong to their respective test sets.}
        \label{tab:retrieval}
        \centering
        \resizebox{\linewidth}{!}{
    \begin{tabular}{lcrrrrrr}
        \toprule
        && \multicolumn{6}{c}{Recall@$k$ (\%) $\uparrow$} \\
        \cmidrule(lr){3-8}
        Method & $k$ & \mlp/\tri & \mlp/\hash & \tri/\mlp & \tri/\hash & \hash/\mlp & \hash/\tri \\
        \midrule
        \nftovec{} & \multirow{2}{*}{--} & -- & -- & -- & -- & -- & -- \\
        \citeauthor{cardace2024neural} && -- & -- & -- & -- & -- & -- \\
        \cmidrule(lr){1-8}
        \textsc{Random} & \multirow{3}{*}{1} & 0.03 & 0.03 & 0.03 & 0.03 & 0.03 & 0.03 \\
        \lr{} (ours) && 1.80 & 0.43 & 4.48 & 1.67 & 1.06 & 0.38 \\
        \lrc{} (ours) && \textbf{30.62} & \textbf{14.77} & \textbf{33.27} & \textbf{13.17} & \textbf{13.45} & \textbf{9.46} \\
        \cmidrule(lr){1-8}
        \textsc{Random} & \multirow{3}{*}{5} & 0.13 & 0.13 & 0.13 & 0.13 & 0.13 & 0.13 \\
        \lr{} (ours) && 5.28 & 1.82 & 13.30 & 6.12 & 3.84 & 1.37 \\
        \lrc{} (ours) && \textbf{59.37} & \textbf{37.98} & \textbf{61.24} & \textbf{33.40} & \textbf{32.09} & \textbf{25.13} \\
        \cmidrule(lr){1-8}
        \textsc{Random} & \multirow{3}{*}{10} & 0.25 & 0.25 & 0.25 & 0.25 & 0.25 & 0.25 \\
        \lr{} (ours) && 8.19 & 3.08 & 19.24 & 9.23 & 6.09 & 2.68 \\
        \lrc{} (ours) && \textbf{72.67} & \textbf{51.25} & \textbf{72.62} & \textbf{45.01} & \textbf{42.83} & \textbf{36.31} \\
        \bottomrule
    \end{tabular}}
    \end{minipage}
    \hfill
    \begin{minipage}[t]{0.51\linewidth}
        \setlength{\tabcolsep}{2.1pt}
        \caption{\textbf{\nerf{} retrieval (Objaverse generalization).} The encoder is trained on \all. Query/gallery combinations belong to their respective test sets.}
        \label{tab:retrieval-obja}
        \centering
        \resizebox{\linewidth}{!}{
    \begin{tabular}{lcrrrrrr}
        \toprule
        && \multicolumn{6}{c}{Recall@$k$ (\%) $\uparrow$} \\
        \cmidrule{3-8}
        Method & $k$ & \mlpob/\triob & \mlpob/\hashob & \triob/\mlpob & \triob/\hashob & \hashob/\mlpob & \hashob/\triob \\
        \midrule
        \nftovec{} & \multirow{2}{*}{--} & -- & -- & -- & -- & -- & -- \\
        \citeauthor{cardace2024neural} && -- & -- & -- & -- & -- & -- \\
        \cmidrule{1-8}
        \textsc{Random} & \multirow{3}{*}{1} & 0.05 & 0.05 & 0.05 & 0.05 & 0.05 & 0.05 \\
        \lr{} (ours) && 2.60 & 0.48 & 2.71 & 0.48 & 0.74 & 0.32 \\
        \lrc{} (ours) && \textbf{9.82} & \textbf{6.10} & \textbf{8.33} & \textbf{1.75} & \textbf{6.74} & \textbf{2.92} \\
        \cmidrule{1-8}
        \textsc{Random} & \multirow{3}{*}{5} & 0.27 & 0.27 & 0.27 & 0.27 & 0.27 & 0.27 \\
        \lr{} (ours) && 6.37 & 1.11 & 7.27 & 1.91 & 2.92 & 0.69 \\
        \lrc{} (ours) && \textbf{23.67} & \textbf{17.94} & \textbf{20.70} & \textbf{5.73} & \textbf{19.59} & \textbf{8.49} \\
        \cmidrule{1-8}
        \textsc{Random} & \multirow{3}{*}{10} & 0.53 & 0.53 & 0.53 & 0.53 & 0.53 & 0.53 \\
        \lr{} (ours) && 9.66 & 2.39 & 10.99 & 3.24 & 4.88 & 1.80 \\
        \lrc{} (ours) && \textbf{33.65} & \textbf{26.38} & \textbf{27.55} & \textbf{9.50} & \textbf{28.13} & \textbf{12.74} \\
        \bottomrule
    \end{tabular}}
    \end{minipage}
\end{table}

\begin{figure}[t]
    \centering
    \includegraphics[width=\linewidth]{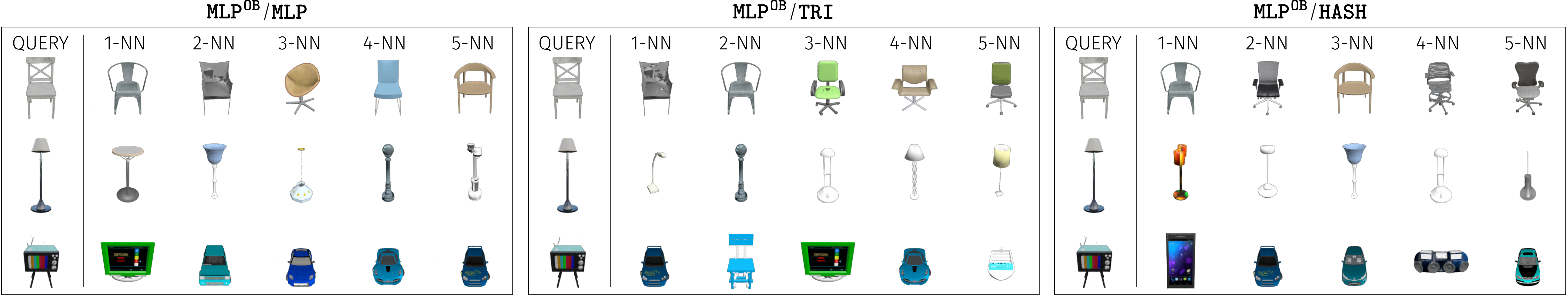}
    \caption{\textbf{Across-datasets \nerf{} retrieval \normalfont{(\lrc{}).}} Query from the test set of \mlpob{} \citep{amaduzzi2025scaling}, gallery from the test set of \mlp{}, \tri{}, or \hash{}.}
    \label{fig:retrieval}
\end{figure}

\subsection{\nerf{} Retrieval}
\label{sec:retrieval}

The embeddings produced by the trained encoder can also be used to perform retrieval tasks via $k$-nearest neighbor search, as outlined in \cref{fig:teaser}. In particular, we define \emph{instance-level} retrieval as follows: given a \nerf{} embedding of a given object (a.k.a.\ the \emph{query}) and a \emph{gallery} of \nerf{} embeddings, the goal is to find the embedding in the gallery that represents the same object as the query but encodes a different \nerf{} architecture.
We evaluate the performance on this task with the recall@$k$ metric \citep{wang2017deep}, defined as the percentage of queries whose $k$ nearest neighbors contain the \nerf{} representing the same object as the query. \cref{tab:retrieval} shows the retrieval results for various query/gallery test-set combinations; the recall@$k$ obtained by randomly shuffling the gallery is shown for reference (\textsc{Random} rows).
Although both \lr{} and \lrc{} significantly outperform the \textsc{Random} baseline, the superior performance of \lrc{} confirms that the contrastive objective favors a more architecture-agnostic latent space. Indeed, \cref{fig:retrieval-full-1,fig:retrieval-full-2} display qualitatively that the latent space organization learned by \lrc{} captures the \nerf{} similarity in color and shape independently of the underlying architecture.

\textbf{Generalization on Objaverse.} To assess the generalization capabilities of our framework, we leverage three additional datasets: \mlpob{}, \triob{}, and \hashob{}. With \mlpob{} we denote a dataset of MLP-based \nerf{}s trained on Objaverse \citep{deitke2023objaverse} recently released by \citet{amaduzzi2025scaling}, whereas \triob{} and \hashob{} are datasets, created by us, of tri-planar and hash-based \nerf{}s trained on the test split of Objaverse defined by \ama{} Specifically, \nerf{}s in \mlpob{}, \triob{}, and \hashob{} have the same architecture as those in \mlp{}, \tri{}, and \hash{}, respectively (see \cref{tab:arch}). \cref{tab:retrieval-obja} mirrors the experiment in \cref{tab:retrieval}, with the notable difference that the query and gallery contain \nerf{}s trained on a different dataset (i.e.\ Objaverse) than the one the model has seen during training (i.e.\ \shapenet{} \citep{xu2019disn}). Although this leads, unsurprisingly, to an overall drop in performance compared to \cref{tab:retrieval}, it is worth noting, once again, that both \lr{} and \lrc{} outperform the \textsc{Random} baseline, and the contrastive loss yields superior results. \cref{fig:retrieval} shows instead a qualitative retrieval experiment where the query is an MLP-based \nerf{} trained on Objaverse (i.e.\ \mlpob{}) while the gallery contains \nerf{}s trained on \shapenet{}, hence the query object is not present in the gallery. Remarkably, for queries that resemble objects featured in \shapenet{}, our method retrieves \nerf{}s that belong to the same class as the query (rows 1--2 of \cref{fig:retrieval}), with some failure cases occurring with more unusual queries (row 3). An extended version of these results is shown in \cref{fig:retrieval-objaverse-full}.

\subsection{\nerf{} Captioning and Q\&A}
\label{sec:language}

\citet{amaduzzi2024llana,amaduzzi2025scaling} have previously shown that language tasks on \nerf{}s benefit from the holistic understanding of 3D objects captured by \nerf{} weights, leading their weight-processing framework (dubbed \llana{}) to outperform methods ingesting rendered views as input in \nerf{} captioning and \nerf{} Q\&A tasks. To achieve these results, \llana{} uses the embeddings produced by the trained \nftovec{} encoder to condition an LLM belonging to the Llama 2 family \citep{touvron2023llama}.
To assess the effectiveness of our learned embeddings in downstream language tasks, we replaced the \nftovec{} encoder in \llana{}'s pipeline with our GMN encoder, trained the resulting framework on textual descriptions from the ShapeNeRF–Text dataset \citep{amaduzzi2024llana}, and evaluated it on brief captioning, detailed captioning, and single-round Q\&A.
In the multi-architecture setting, our framework is trained on \all{} (as usual), and the \all{} embeddings produced by our encoder are used to train the LLM. Since the \nftovec{} encoder can process only MLP-based NeRFs, the original LLaNA framework cannot be applied in the multi-architecture setting. In the single-architecture setting, instead, our framework is trained on \mlp{}, and the \mlp{} embeddings produced by our encoder are used to train the LLM. In this setting, our results are compared with the original \llana{} framework, which is also trained on \mlp{}.
Multi-architecture results are shown in \cref{tab:caption-brief-multi,tab:caption-detailed-multi,tab:q&a-multi}, whereas single-architecture results are shown in \cref{tab:caption-brief-single,tab:caption-detailed-single,tab:q&a-single}. As in \citet{amaduzzi2024llana,amaduzzi2025scaling}, the reported metrics are a measure of the similarity between the global embeddings of the generated and ground-truth
texts, where such embeddings are produced by pre-trained Sentence-BERT \citep{reimers2019sentence} and SimCSE \citep{gao2021simcse} encoders, respectively. Extended results including traditional n-gram-based metrics are shown in \cref{app:language}. Two main conclusions can be drawn from \cref{tab:caption-brief-multi,tab:caption-detailed-multi,tab:q&a-multi,tab:caption-brief-single,tab:caption-detailed-single,tab:q&a-single}: in the multi-architecture setting, our method is remarkably robust to the choice of test-time NeRF architecture; when instead restricted to MLP-based \nerf{}s only, our method achieves results comparable to those of the original \llana{} framework. These results further confirm the architecture-agnostic nature of our embeddings and their effectiveness in downstream tasks, even as complex as those involving multimodal \nerf{}--language understanding.

\begin{table}[t]
    \centering
    \begin{minipage}[t]{0.32\linewidth}
        \caption{\textbf{\nerf{} brief captioning (multi-architecture)}.}
        \label{tab:caption-brief-multi}
        \centering
        \resizebox{\linewidth}{!}{
        \begin{tabular}{lccc}
            \toprule
            Method & Test Dataset & S-BERT $\uparrow$ & SimCSE $\uparrow$ \\
            \midrule
            \llana{} & -- & -- & -- \\
            \cmidrule(lr){1-4}
            \multirow{3}{*}{\lrc{} (ours)} & \mlp{} & \textbf{67.1} & \textbf{68.9} \\
            & \tri{} & 66.4 & 67.8 \\
            & \hash{} & 66.3 & 67.9 \\
            \bottomrule
        \end{tabular}}
        \end{minipage}
    \hfill
    \begin{minipage}[t]{0.32\linewidth}
        \caption{\textbf{\nerf{} detailed captioning (multi-architecture)}.}
        \label{tab:caption-detailed-multi}
        \centering
        \resizebox{\linewidth}{!}{
        \begin{tabular}{lccc}
            \toprule
            Method & Test Dataset & S-BERT $\uparrow$ & SimCSE $\uparrow$ \\
            \midrule
            \llana{} & -- & -- & -- \\
            \cmidrule(lr){1-4}
            \multirow{3}{*}{\lrc{} (ours)} & \mlp{} & \textbf{73.3} & \textbf{75.1}  \\
            & \tri{} & 72.2 & 73.8 \\
            & \hash{} & 72.4 & 74.0 \\
            \bottomrule
        \end{tabular}}
    \end{minipage}
    \hfill
    \begin{minipage}[t]{0.32\linewidth}
        \caption{\textbf{\nerf{} single-round Q\&A (multi-architecture)}.}
        \label{tab:q&a-multi}
        \centering
        \resizebox{\linewidth}{!}{
        \begin{tabular}{lccc}
            \toprule
            Method & Test Dataset & S-BERT $\uparrow$ & SimCSE $\uparrow$ \\
            \midrule
            \llana{} & -- & -- & -- \\
            \cmidrule(lr){1-4}
            \multirow{3}{*}{\lrc{} (ours)} & \mlp{} & \textbf{80.9} & \textbf{81.4}  \\
            & \tri{} & 80.7 & 81.3 \\
            & \hash{} & 80.6 & 81.3 \\
            \bottomrule
        \end{tabular}}
    \end{minipage}
    
    \hfill
    \begin{minipage}[t]{0.32\linewidth}
        \caption{\textbf{\nerf{} brief captioning (single-architecture)}.}
        \label{tab:caption-brief-single}
        \centering
        \resizebox{\linewidth}{!}{
        \begin{tabular}{lcc}
            \toprule
            Method & S-BERT $\uparrow$ & SimCSE $\uparrow$ \\
            \midrule
            \llana{} & \textbf{68.6} & \textbf{70.5} \\
            \lr{} (ours) & 68.3 & 70.2 \\
            \bottomrule
        \end{tabular}}
    \end{minipage}
    \hfill
    \begin{minipage}[t]{0.32\linewidth}
        \caption{\textbf{\nerf{} detailed captioning (single-architecture)}.}
        \label{tab:caption-detailed-single}
        \centering
        \resizebox{\linewidth}{!}{
        \begin{tabular}{lcc}
            \toprule
            Method & S-BERT $\uparrow$ & SimCSE $\uparrow$ \\
            \midrule
            \llana{} & \textbf{77.4} & \textbf{79.8} \\
            \lr{} (ours) & 74.1 & 76.0 \\
            \bottomrule
        \end{tabular}}
    \end{minipage}
    \hfill
    \begin{minipage}[t]{0.32\linewidth}
        \caption{\textbf{\nerf{} single-round Q\&A (single-architecture)}.}
        \label{tab:q&a-single}
        \centering
        \resizebox{\linewidth}{!}{
        \begin{tabular}{lcc}
            \toprule
            Method & S-BERT $\uparrow$ & SimCSE $\uparrow$ \\
            \midrule
            \llana{} & \textbf{81.0} & \textbf{81.6} \\
            \lr{} (ours) & \textbf{81.0} & \textbf{81.6} \\
            \bottomrule
        \end{tabular}}
    \end{minipage}
\end{table}

\section{Conclusions}
\label{sec:conclusions}

In this paper, we presented the first framework that performs downstream tasks on \nerf{} weights with diverse architectures. Our method, based on a Graph Meta-Network encoder ingesting the \nerf{} parameter graph, can
\begin{enumerate*}[label=(\roman*)]
    \item handle input \nerf{}s parameterized by MLPs, tri-planes, and, for the first time, multi-resolution hash tables;
    \item process, at inference time, variations of architectures seen during training;
    \item provide state-of-the-art performance in the single-architecture scenario.
\end{enumerate*}
Furthermore, we investigated the interplay between rendering and contrastive training objectives and showed that they serve complementary purposes, by favoring either class-level separability or invariance to the \nerf{} architecture. 
The main limitation of our study lies in its experimental evaluation, which primarily focuses on \nerf{}s that, although parameterized by 13 total architectures, are trained on a single dataset, i.e.\ \shapenet{} \citep{xu2019disn}, with the notable exception of two generalization experiments on Objaverse \citep{deitke2023objaverse}. Extending our training procedure on Objaverse itself or similar large-scale \nerf{} datasets would further validate the effectiveness of our approach, and we plan to follow this research path in future work. Eventually, our methodology could be scaled up to become the first foundational model for \nerf{} weight space processing.

\newpage
\bibliography{main}
\bibliographystyle{iclr2026_conference}

\newpage
\appendix
\crefalias{section}{appendix}

\section{Additional Parameter Graph Details}
\label{app:graph}

\begin{figure}
    \centering
    \includegraphics[width=\linewidth]{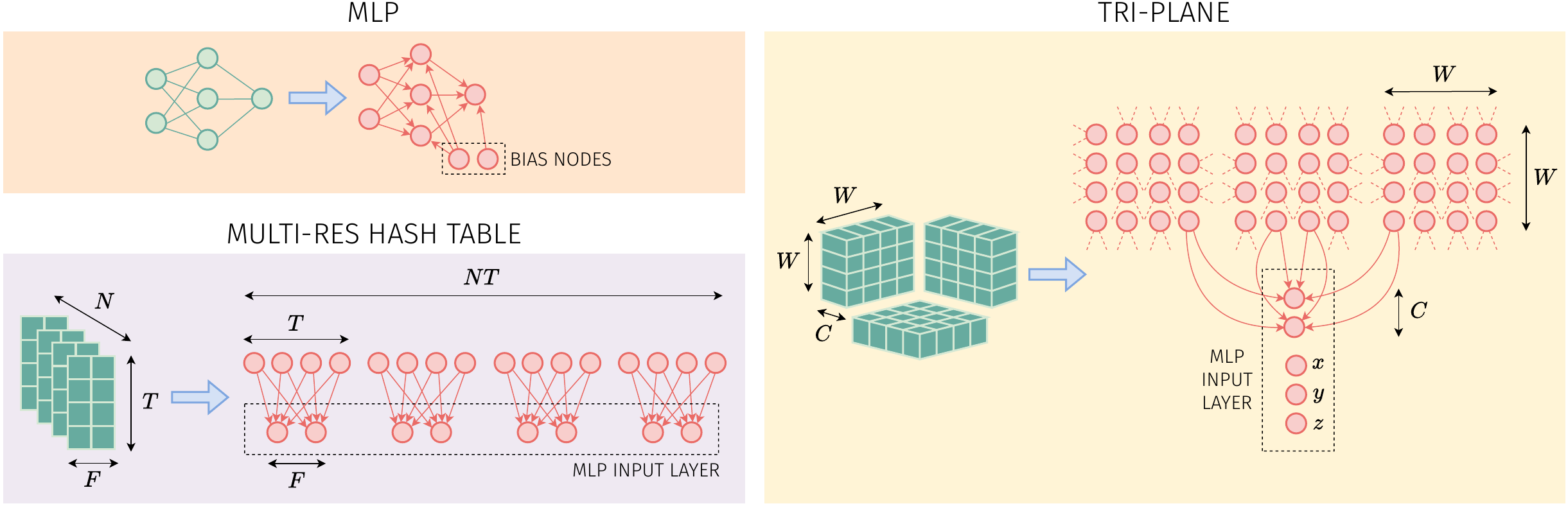}
    \caption{\textbf{Parameter graph conversion.} \textbf{Top left:} parameter graph representation of an MLP, proposed by \citet{lim2024graph}. \textbf{Right:} parameter graph representation of a tri-plane, proposed by \citet{lim2024graph}. Dotted edges should be connected to the $C$ channel nodes, but are not fully drawn for better visual clarity. \textbf{Bottom left:} our parameter graph representation of a multi-resolution hash table.}
    \label{fig:graph-app}
\end{figure}

The parameter graph conversion of an MLP, a tri-plane, and a multi-resolution hash table is depicted in \cref{fig:graph-app}, with additional details compared to \cref{fig:method} (left).

\textbf{MLP.} The parameter graph of a linear layer (proposed by \citet{lim2024graph}) coincides with its computation graph, except for biases, which are modeled by including an additional node for each (non-input) layer, connected to every other neuron in the layer with an edge. As a result, the parameter graph of an MLP is identical to its computation graph, with the addition of the aforementioned bias nodes. Each node stores three integer values: layer index, neuron index within the layer, and node type (\emph{neuron} or \emph{bias}). Each edge stores the corresponding weight or bias and two integers, i.e.\ layer index and edge type (\emph{linear layer weight} or \emph{linear layer bias}).

\textbf{Tri-plane.} The parameter graph of a tri-plane (proposed by \citet{lim2024graph}) contains one node for each spatial location and one node for each feature channel, where each spatial node is connected via an edge to every channel node; the channel nodes, together with one node for each input coordinate (more precisely, one node for each entry in their frequency encoding; see \cref{app:details}) become the first layer of the MLP parameter graph that follows. Each node stores three integer values: layer index, neuron index within the layer, and node type (\emph{neuron} for coordinate nodes or \emph{tri-planar} for spatial and channel nodes). Each edge stores the corresponding learnable parameter (a.k.a.\ feature) and five additional values: layer index (integer), edge type (\emph{tri-planar}, integer), and three $(i,j,k)$ values (evenly spaced floats $\in[-1,1]$) denoting the spatial position of the parameter within the tri-plane (with $i=0$ on the $yz$ plane, $j=0$ in the $xz$ plane, and $k=0$ on the $xy$ plane).

\textbf{Hash table.} As mentioned in \cref{sec:method}, we are the first to propose a parameter graph construction for multi-resolution hash tables. A single hash table is modeled with a node for each table entry and a node for each feature vector dimension, where each entry node is connected to each feature node via an edge. This subgraph construction is repeated for each hash table, i.e.\ for each resolution level. Finally, the concatenation of all feature nodes becomes the first layer of the MLP parameter graph that follows. Each node stores three integer values: layer index, neuron index within the layer, and node type (\emph{hash table}). Each edge stores the corresponding learnable parameter (a.k.a.\ feature) and four additional integers: layer index, edge type (\emph{hash table}), table index, and entry index within the table. Consider a multi-resolution hash table with $N$ levels, max table size $T$, $F$ features per entry, and max resolution $R$. Our parameter graph construction requires $N(T+F)$ nodes and $NTF$ edges. Conversely, if we drew an analogy with the tri-planar graph representation and explicitly modeled the underlying voxel grids with one node for each spatial location, it would result in a graph with $\mathcal{O}(N(R^3+F))$ nodes and $\mathcal{O}(NR^3F)$ edges. Since the purpose of hash grids is to reduce the memory footprint by mapping multiple voxel grid locations to the same table entry via a hashing function, $T$ should always be much smaller than $R^3$ (e.g.\ \nerf{}s in \hash{} have $T=2^{12}$ vs $R^3=2^{21}$), hence $N(T+F)\ll N(R^3+F)$ and $NTF\ll NR^3F$. Therefore, the motivation behind our choice of parameter graph representation for hash tables lies in its higher memory efficiency compared to its voxel-grid-based alternative.

\begin{table}
    \caption{\textbf{Dataset architectures.} Extended version of \cref{tab:arch}. Architectural hyperparameters of \nerf{} datasets featured in our experiments, including those used at inference time only (\emph{unseen}).}
    \label{tab:arch-app}
    \centering
    \resizebox{\linewidth}{!}{
    \begin{tabular}{lrrrrrrrrrrrrr}
        \toprule
        & \multicolumn{3}{c}{Training} & \multicolumn{10}{c}{Unseen (ours)} \\
        \cmidrule(lr){2-4} \cmidrule(lr){5-14}
        & \mlp{} & \tri{} & \hash{} & \mlpl{} & \mlph{} & \tril{} & \trih{} & \triw{} & \tric{} & \hashl{} & \hashh{} & \hashn{} & \hasht{} \\
        \midrule
        Frequency Encoding & \cmark & \cmark & \xmark & \cmark & \cmark & \cmark & \cmark & \cmark & \cmark & \xmark & \xmark & \xmark & \xmark \\
        MLP Activation & ReLU & Sine & ReLU & ReLU & ReLU & Sine & Sine & Sine & Sine & ReLU & ReLU & ReLU & ReLU \\
        MLP Hidden Layers & 3 & 3 & 3 & 2 & 3 & 2 & 3 & 3 & 3 & 2 & 3 & 3 & 3 \\
        MLP Hidden Dimension & 64 & 64 & 64 & 64 & 32 & 64 & 32 & 64 & 64 & 64 & 32 & 64 & 64 \\
        Tri-plane Resolution & -- & 32 & -- & -- & -- & 32 & 32 & 16 & 32 & -- & -- & -- & -- \\
        Tri-plane Channels & -- & 16 & -- & -- & -- & 16 & 16 & 16 & 8 & -- & -- & -- & -- \\
        Hash Table Levels & -- & -- & 4 & -- & -- & -- & -- & -- & -- & 4 & 4 & 3 & 4 \\
        Hash Table Size ($\log_2$) & -- & -- & 12 & -- & -- & -- & -- & -- & -- & 12 & 12 & 12 & 11 \\
        Hash Table Features per Entry & -- & -- & 2 & -- & -- & -- & -- & -- & -- & 2 & 2 & 2 & 2 \\
        \bottomrule
    \end{tabular}}
\end{table}

\section{Additional Experimental Details}
\label{app:details}

\textbf{\nerf{} datasets.} As outlined in \cref{sec:experiments}, the datasets used throughout our experiments mostly consist of \nerf{}s trained on \shapenet{} \citep{xu2019disn}. \shapenet{} is a dataset featuring RGB images and class labels of synthetic objects belonging to 13 classes of the ShapeNetCore dataset \citep{shapenet}: \emph{airplane}, \emph{bench}, \emph{cabinet}, \emph{car}, \emph{chair}, \emph{display}, \emph{lamp}, \emph{speaker}, \emph{rifle}, \emph{sofa}, \emph{table}, \emph{phone}, and \emph{watercraft}. For each object, the dataset contains 36 $224\times224$ images. We follow \citet{ramirez2024deep}'s train-val-test split without augmentations, consisting of 31744 \nerf{}s used for training, 3961 for validation, and 3961 for testing.
\cref{tab:arch-app} summarizes the \nerf{} architectures featured in our experiments. In addition to the dataset descriptions provided in \cref{sec:experiments}, we note here that:
{\setlist{nosep}
\begin{itemize}[leftmargin=*]
    \item In \mlp{} and \tri{}, input $(x,y,z)$ coordinates go through a frequency encoding \citep{mildenhall2020nerf} before being concatenated to the tri-plane or hash table output that gets fed to the MLP, as this choice was made by the original creators of the two datasets, i.e. \citet{ramirez2024deep} and \citet{cardace2024neural}, respectively. Conversely, the MLP in \hash{} takes as input the output of the multi-resolution hash table by itself, without (encoded) input coordinates being concatenated to it; we choose to do so in order to be faithful to the original multi-resolution hash table architecture proposed by \citet{muller2022instant}. The presence or absence of concatenated encoded coordinates in a dataset used for training is then reflected in all the unseen variations of that architecture family (e.g.\ \mlpl{} uses frequency encoding, whereas \hashl{} does not).
    \item MLP layers are intertwined with ReLU activation functions in \mlp{} and \hash{}, whereas \tri{} uses sine activation functions \citep{sitzmann2020implicit}. The activation choices for \mlp{} and \tri{} were made by their original creators \citep{ramirez2024deep,cardace2024neural}, whereas we chose ReLU activations for \hash{} to remain faithful to the original hash table architecture \citep{muller2022instant}. The activation choice in a dataset used for training is then reflected in all the unseen variations of that architecture family (e.g.\ \mlpl{} uses ReLU, whereas \tril{} uses sine).
    \item All \nerf{}s approximate a function from input coordinates to predicted color and density values, i.e.\ $(x,y,z)\mapsto(r,g,b,\sigma)$. This follows the simplified \nerf{} formulation proposed by \citet{mildenhall2020nerf}, the alternative being adding a viewing direction $(\theta,\phi)$ as input and letting the \nerf{} approximate a function $(x,y,z,\theta,\phi)\mapsto(r,g,b,\sigma)$. The formulation with viewing direction is usually modeled by two MLPs: a \emph{density MLP} taking as input $(x,y,z)$ (or, more precisely, an encoding of $(x,y,z)$) and producing as output $\sigma$ together with an additional vector $\mathbf{h}$ of features, and a \emph{color MLP} taking as input the concatenation of $\mathbf{h}$ and $(\theta,\phi)$ and predicting $(r,g,b)$ as output. Both \citet{ramirez2024deep} and \citet{cardace2024neural} adopt the simplified $(x,y,z)\mapsto(r,g,b,\sigma)$ formulation modeled by a single MLP when creating their \nerf{} datasets (i.e.\ \mlp{} and \tri{}, respectively). We choose to follow the same formulation for \hash{} and the datasets of unseen architectures, both to keep the datasets consistent and to reduce the complexity of the parameter graph going into the encoder.
\end{itemize}}
Furthermore, the experiments in \cref{sec:retrieval} leverage three additional test sets, dubbed \mlpob{}, \triob{}, and \hashob{}. \mlpob{} is a dataset of MLP-based \nerf{}s trained on Objaverse \citep{deitke2023objaverse}, a large-scale dataset of 3D objects. \mlpob{} was recently released by \citet{amaduzzi2025scaling}, whereas we created \triob{} and \hashob{} ourselves by training tri-planar and hash-based \nerf{}s on the Objaverse test split used in \citet{amaduzzi2025scaling}. More precisely, our test split is the union of the two test splits featured in \citet{amaduzzi2025scaling}'s benchmark, i.e.\ the one adopted by PointLLM \citep{xu2024pointllm} and the one adopted by GPT4Point \citep{qi2024gpt4point}. The resulting test set contains 1884 \nerf{}s. \nerf{}s in \mlpob{}, \triob{}, and \hashob{} have the same architecture as those in \mlp{}, \tri{}, and \hash{}, respectively.

\textbf{Encoder architecture.} Our encoder is a slight re-adaptation of the graph meta-network used in \citet{lim2024graph} in their experiment called \emph{predicting accuracy for varying architectures}. It has the same network hyperparameters: hidden dimension 128, 4 layers, a pre-encoder, 2 readout layers, and uses directed edges. However, unlike \citet{lim2024graph}, before average-pooling the edge features, it maps them to vectors of size 1024 through a single linear layer, resulting in a final embedding of size 1024 as in \nftovec{} \citep{ramirez2024deep}.

\textbf{Decoder architecture.} We use the \nftovec{} decoder in its original formulation \citep{ramirez2024deep}. The input coordinates $(x,y,z)$ are mapped to a vector of size 144 through a frequency encoding \citep{mildenhall2020nerf} and concatenated with the 1024-dimensional embedding produced by the encoder. The resulting vector of size 1168 becomes the input of the decoder, which consists of 4 hidden layers with dimension 1024 intertwined with ReLU activations, with a skip connection that maps the input vector to a vector of size 1024 via a single linear layer + ReLU and sums it to the output of the second hidden layer. Finally, the decoder outputs the four $(r,g,b,\sigma)$ predicted radiance field values.

\textbf{Training.} Our encoder-decoder framework is trained end-to-end for 250 epochs with batch size 8, AdamW optimizer \citep{loshchilov2018decoupled}, one-cycle learning rate scheduler \citep{onecycle}, maximum learning rate \num{1e-4} and weight decay \num{1e-2}. These training hyperparameters are the same for \lr{}, \lrc{}, and \lc{}. In \cref{eq:rend}, foreground and background pixels are weighted by 0.8 and 0.2, respectively, as done in \citet{ramirez2024deep}. In \cref{eq:contrastive}, $t$ and $b$ are initialized to $10$ and $-10$, respectively, as done in \citet{zhai2023sigmoid}. In \cref{eq:combined}, $\lambda$ is set to \num{2e-2}, as we experimentally observed that this choice leads to \lr{} and \lc{} values with the same order of magnitude.

\textbf{Classification.} In our classification experiments of \cref{sec:classification}, the classifier is a concatenation of 3 (linear $\rightarrow$ batch-norm $\rightarrow$ ReLU $\rightarrow$ dropout) blocks, where the hidden dimensions of the linear layers are 1024, 512, and 256, and a final linear layer at the end computes the class logits. The classifier is trained via cross-entropy loss for 150 epochs with batch size 256, AdamW optimizer \citep{loshchilov2018decoupled}, one-cycle learning rate scheduler \citep{onecycle}, maximum learning rate \num{1e-4}, and weight decay \num{1e-2}. These same network and training hyperparameters are used in the classification experiments of \citet{ramirez2024deep}.

\begin{table}
    \centering
    \caption{\textbf{\nerf{} classification (multi-architecture).} Extended version of \cref{tab:classification-multi} with added \lc{} results. The encoder is trained on \all{}; the classifier is trained on the datasets in column 2 and tested on those in columns 3--5.}
    \label{tab:classification-multi-app}
    \centering
    \resizebox{0.7\linewidth}{!}{
    \begin{tabular}{lcrrr}
        \toprule
        && \multicolumn{3}{c}{Accuracy (\%) $\uparrow$} \\
        \cmidrule(lr){3-5}
        Method & Classifier Training Set & \mlp & \tri & \hash \\
        \midrule
        \nftovec{} \citep{ramirez2024deep} & \multirow{2}{*}{--} & -- & -- & -- \\
        \citet{cardace2024neural} && -- & -- & -- \\
        \cmidrule(lr){1-5}
        \lr{} (ours) & \multirow{3}{*}{\all} & \cellcolor{red!15} \textbf{93.6} & \cellcolor{red!15} \textbf{94.0} & \cellcolor{red!15} \textbf{92.4} \\
        \lrc{} (ours) && \cellcolor{red!15} 90.7 & \cellcolor{red!15} 90.6 & \cellcolor{red!15} 90.0 \\
        \lc{} (ours) && \cellcolor{red!15} 82.5 & \cellcolor{red!15} 77.8 & \cellcolor{red!15} 64.0 \\
        \cmidrule(lr){1-5}
        \lr{} (ours) & \multirow{3}{*}{\mlp} & \cellcolor{red!15} \textbf{93.8} & 25.3 & 19.3 \\
        \lrc{} (ours) && \cellcolor{red!15} 91.5 & 58.6 & \textbf{56.6} \\
        \lc{} (ours) && \cellcolor{red!15} 83.2 & \textbf{60.9} & 38.6 \\
        \cmidrule(lr){1-5}
        \lr{} (ours) & \multirow{3}{*}{\tri} & 11.4 & \cellcolor{red!15} \textbf{93.8} & 9.3 \\
        \lrc{} (ours) && \textbf{77.8} & \cellcolor{red!15} 91.0 & \textbf{66.5} \\
        \lc{} (ours) && 70.4 & \cellcolor{red!15} 79.0 & 47.6 \\
        \cmidrule(lr){1-5}
        \lr{} (ours) & \multirow{3}{*}{\hash} & 13.8 & 35.7 & \cellcolor{red!15} \textbf{92.7} \\
        \lrc{} (ours) && 54.1 & 35.5 & \cellcolor{red!15} 90.6 \\
        \lc{} (ours) && \textbf{65.4} & \textbf{63.7} & \cellcolor{red!15} 64.6 \\
        \bottomrule
    \end{tabular}}
\end{table}

\begin{table}
    \caption{\textbf{\nerf{} classification of unseen architectures (multi-architecture).} Extended version of \cref{tab:classification-multi-unseen} with added \lc{} results. The encoder is trained on \all{}; the classifier is trained on the datasets in column 2 and tested on those in columns 3--12, containing \nerf{} architectures unseen at training time.}
    \label{tab:classification-multi-unseen-app}
    \centering
    \resizebox{\linewidth}{!}{
    \begin{tabular}{lcrrrrrrrrrr}
        \toprule
        && \multicolumn{10}{c}{Accuracy (\%) $\uparrow$} \\
        \cmidrule(lr){3-12}
         \multirow{2}{*}{Method} & Classifier & \multirow{2}{*}{\mlpl} & \multirow{2}{*}{\mlph} & \multirow{2}{*}{\tril} & \multirow{2}{*}{\trih} & \multirow{2}{*}{\triw} & \multirow{2}{*}{\tric} & \multirow{2}{*}{\hashl} & \multirow{2}{*}{\hashh} & \multirow{2}{*}{\hashn} & \multirow{2}{*}{\hasht}\\
        & Training Set &&&&&& \\
        \midrule
        \nftovec{} \citep{ramirez2024deep} & -- & -- & -- & -- & -- & -- & -- & -- & -- & -- & -- \\
        \citet{cardace2024neural} & -- & -- & -- & -- & -- & -- & -- & -- & -- & -- & -- \\
        \cmidrule(lr){1-12}
        \lr{} (ours) & \multirow{3}{*}{\all} & \cellcolor{orange!15} \textbf{91.3} & \cellcolor{orange!15} \textbf{87.4} & \cellcolor{orange!15} \textbf{93.2} & \cellcolor{orange!15} \textbf{88.3} & \cellcolor{orange!15} 24.8 & \cellcolor{orange!15} \textbf{69.4} & \cellcolor{orange!15} \textbf{91.9} & \cellcolor{orange!15} \textbf{91.2} & \cellcolor{orange!15} \textbf{88.3} & \cellcolor{orange!15} 24.5 \\
        \lrc{} (ours) && \cellcolor{orange!15} 85.9 & \cellcolor{orange!15} 83.8 & \cellcolor{orange!15} 87.0 & \cellcolor{orange!15} 84.1 & \cellcolor{orange!15} \textbf{72.1} & \cellcolor{orange!15} 30.8 & \cellcolor{orange!15} 89.2 & \cellcolor{orange!15} 87.4 & \cellcolor{orange!15} 86.8 & \cellcolor{orange!15} \textbf{27.8} \\
        \lc{} (ours) && \cellcolor{orange!15} 67.3 & \cellcolor{orange!15} 47.7 & \cellcolor{orange!15} 63.5 & \cellcolor{orange!15} 43.3 & \cellcolor{orange!15} 60.4 & \cellcolor{orange!15} 4.6 & \cellcolor{orange!15} 51.9 & \cellcolor{orange!15} 43.5 & \cellcolor{orange!15} 55.9 & \cellcolor{orange!15} 27.7 \\
        \cmidrule(lr){1-12}
        \lr{} (ours) &\multirow{3}{*}{\mlp} & \cellcolor{orange!15} \textbf{91.3} & \cellcolor{orange!15} \textbf{86.1} & 22.6 & 23.2 & 7.9 & \textbf{21.0} & 19.7 & 20.6 & 21.8 & 7.6 \\
        \lrc{} (ours) && \cellcolor{orange!15} 86.6 & \cellcolor{orange!15} 81.3 & \textbf{63.3} & \textbf{43.4} & \textbf{59.2} & 13.1 & \textbf{51.5} & \textbf{50.8} & \textbf{55.0} & 24.8 \\
        \lc{} (ours) && \cellcolor{orange!15} 69.6 & \cellcolor{orange!15} 52.6 & 59.7 & 42.0 & 55.2 & 6.0 & 33.0 & 23.3 & 39.4 & \textbf{29.5} \\
        \cmidrule(lr){1-12}
        \lr{} (ours) & \multirow{3}{*}{\tri} & 11.9 & 10.6 & \cellcolor{orange!15} \textbf{92.1} & \cellcolor{orange!15} \textbf{84.8} & \cellcolor{orange!15} 32.1 & \cellcolor{orange!15} \textbf{43.6} & 9.2 & 8.8 & 15.0 & 5.2 \\
        \lrc{} (ours) && \textbf{60.7} & \textbf{58.0} & \cellcolor{orange!15} 86.9 & \cellcolor{orange!15} 83.2 & \cellcolor{orange!15} \textbf{63.9} & \cellcolor{orange!15} 30.2 & \textbf{59.8} & \textbf{61.5} & \textbf{66.3} & \textbf{28.5} \\
        \lc{} (ours) && 54.6 & 37.5 & \cellcolor{orange!15} 62.8 & \cellcolor{orange!15} 44.9 & \cellcolor{orange!15} 57.3 & \cellcolor{orange!15} 5.8 & 36.2 & 27.9 & 43.1 & 28.4 \\
        \cmidrule(lr){1-12}
        \lr{} (ours) & \multirow{3}{*}{\hash} & 10.7 & 7.4 & 33.9 & \textbf{36.3} & 19.1 & \textbf{23.0} & \cellcolor{orange!15} \textbf{91.6} & \cellcolor{orange!15} \textbf{91.4} & \cellcolor{orange!15} \textbf{87.8} & \cellcolor{orange!15} \textbf{29.2} \\
        \lrc{} (ours) && 47.0 & \textbf{40.4} & 40.6 & 34.0 & \textbf{43.4} & 19.3 & \cellcolor{orange!15} 89.5 & \cellcolor{orange!15} 87.9 & \cellcolor{orange!15} 86.4 & \cellcolor{orange!15} 25.7 \\
        \lc{} (ours) && \textbf{56.8} & 39.1 & \textbf{48.7} & 34.1 & 43.0 & 4.3 & \cellcolor{orange!15} 55.7 & \cellcolor{orange!15} 46.0 & \cellcolor{orange!15} 56.3 & \cellcolor{orange!15} 25.2 \\
        \bottomrule
    \end{tabular}}
\end{table}

\section{\lc{} Classification and Retrieval Results}
\label{app:experiments}

\textbf{\nerf{} classification.} \cref{tab:classification-multi-app,tab:classification-multi-unseen-app} show classification results in the multi-architecture setting, including those produced by \lc{}, which are omitted in \cref{tab:classification-multi,tab:classification-multi-unseen}. Specifically, \cref{tab:classification-multi-app} shows classification results when the classifier $\mathcal{C}$ is tested on architectures seen during training, whereas \cref{tab:classification-multi-unseen-app} shows classification results when $\mathcal{C}$ is tested on architectures belonging to the families seen at training time but with different hyperparameters, which we refer to as \emph{unseen} architectures in \cref{sec:experiments}. In addition to the observations discussed in \cref{sec:classification}, \cref{tab:classification-multi-app,tab:classification-multi-unseen-app} show that \lc{} prevails over \lr{} and \lrc{} in very few cases; this behavior is consistent with the less noticeable separation between class-level clusters produced by \lc{} shown in \cref{fig:tsne} and discussed in \cref{sec:latent}. Notably, \lc{} in the best performing method when $\mathcal{C}$ is trained on \hash{} and tested on \tri{} in \cref{tab:classification-multi-app}, where \lrc{} struggles much more than in all other cases where $\mathcal{C}$ is tested on architectures that are not included in its training set (i.e.\ white cells).

\textbf{\nerf{} retrieval.} \cref{tab:retrieval-app,tab:retrieval-obja-app} show retrieval results including those produced by \lc{}, which are omitted in \cref{tab:retrieval,tab:retrieval-obja}. The recall@$k$ of the \textsc{Random} baseline is computed as
\begin{equation}
    100\cdot\sum_{i=1}^k\frac{1}{|\text{gallery}|} = \frac{100\cdot k}{|\text{gallery}|}
\end{equation}
Since retrieval is an instance-level task, the contrastive loss enables \lc{} to outperform \lr{} for every query/gallery combination, albeit not as effectively as \lrc{}, which consistently yields the best results. In particular, when both the query and the gallery belong to \all{}, \lr{} is unable to perform the task: the latent space organization shown in \cref{fig:tsne} (left) and discussed in \cref{sec:latent} prevents \lr{} from being capable of recognizing the same object represented by \nerf{}s parameterized by different architectures. Furthermore, it is worth noting that the recall@$k$ in the \all{}/\all{} case is lower than in the other query/gallery combinations because the gallery is larger and the probability of missing the target instance is thus higher.

\begin{table}
    \caption{\textbf{\nerf{} retrieval (\shapenet{}).} Extended version of \cref{tab:retrieval} with added \lc{} results. The encoder is trained on \all. Query/gallery combinations belong to their respective test sets.}
    \label{tab:retrieval-app}
    \centering
    \resizebox{\linewidth}{!}{
    \begin{tabular}{lcrrrrrrr}
        \toprule
        && \multicolumn{7}{c}{Recall@$k$ (\%) $\uparrow$} \\
        \cmidrule(lr){3-9}
        Method & $k$ & \all/\all & \mlp/\tri & \mlp/\hash & \tri/\mlp & \tri/\hash & \hash/\mlp & \hash/\tri \\
        \midrule
        \nftovec{} \citep{ramirez2024deep} & \multirow{2}{*}{--} & -- & -- & -- & -- & -- & -- & -- \\
        \citet{cardace2024neural} && -- & -- & -- & -- & -- & -- & -- \\
        \cmidrule(lr){1-9}
        \textsc{Random} & \multirow{4}{*}{1} & 0.01 & 0.03 & 0.03 & 0.03 & 0.03 & 0.03 & 0.03 \\
        \lr{} (ours) && 0.00 & 1.80 & 0.43 & 4.48 & 1.67 & 1.06 & 0.38 \\
        \lrc{} (ours) && \textbf{5.25} & \textbf{30.62} & \textbf{14.77} & \textbf{33.27} & \textbf{13.17} & \textbf{13.45} & \textbf{9.46} \\
        \lc{} (ours) && 2.86 & 16.41 & 6.19 & 15.93 & 4.35 & 5.39 & 3.99 \\
        \cmidrule(lr){1-9}
        \textsc{Random} & \multirow{4}{*}{5} & 0.04 & 0.13 & 0.13 & 0.13 & 0.13 & 0.13 & 0.13 \\
        \lr{} (ours) && 0.00 & 5.28 & 1.82 & 13.30 & 6.12 & 3.84 & 1.37 \\
        \lrc{} (ours) && \textbf{17.72} & \textbf{59.37} & \textbf{37.98} & \textbf{61.24} & \textbf{33.40} & \textbf{32.09} & \textbf{25.13} \\
        \lc{} (ours) && 10.91 & 40.61 & 18.03 & 38.89 & 14.11 & 16.84 & 14.21 \\
        \cmidrule(lr){1-9}
        \textsc{Random} & \multirow{4}{*}{10} & 0.08 & 0.25 & 0.25 & 0.25 & 0.25 & 0.25 & 0.25 \\
        \lr{} (ours) && 0.00 & 8.19 & 3.08 & 19.24 & 9.23 & 6.09 & 2.68 \\
        \lrc{} (ours) && \textbf{27.25} & \textbf{72.67} & \textbf{51.25} & \textbf{72.62} & \textbf{45.01} & \textbf{42.83} & \textbf{36.31} \\
        \lc{} (ours) && 18.03 & 53.75 & 26.93 & 53.63 & 22.05 & 26.30 & 21.95 \\
        \bottomrule
    \end{tabular}}
\end{table}

\begin{table}
    \vspace{-5pt}
    \caption{\textbf{\nerf{} retrieval (Objaverse generalization).} Extended version of \cref{tab:retrieval-obja} with added \lc{} results.. The encoder is trained on \all. Query/gallery combinations belong to their respective test sets.}
    \label{tab:retrieval-obja-app}
    \centering
    \resizebox{\linewidth}{!}{
    \begin{tabular}{lcrrrrrrr}
        \toprule
        && \multicolumn{7}{c}{Recall@$k$ (\%) $\uparrow$} \\
        \cmidrule(lr){3-9}
        Method & $k$ & \all/\all & \mlp/\tri & \mlp/\hash & \tri/\mlp & \tri/\hash & \hash/\mlp & \hash/\tri \\
        \midrule
        \nftovec{} \citep{ramirez2024deep} & \multirow{2}{*}{--} & -- & -- & -- & -- & -- & -- & -- \\
        \citet{cardace2024neural} && -- & -- & -- & -- & -- & -- & -- \\
        \cmidrule(lr){1-9}
        \textsc{Random} & \multirow{4}{*}{1} & 0.02 & 0.05 & 0.05 & 0.05 & 0.05 & 0.05 & 0.05 \\
        \lr{} (ours) && 0.00 & 2.60 & 0.48 & 2.71 & 0.48 & 0.74 & 0.32 \\
        \lrc{} (ours) && \textbf{0.64} & \textbf{9.82} & \textbf{6.10} & \textbf{8.33} & \textbf{1.75} & \textbf{6.74} & \textbf{2.92} \\
        \lc{} (ours) && 0.18 & 3.66 & 2.23 & 3.66 & 0.96 & 3.13 & 1.80 \\
        \cmidrule(lr){1-9}
        \textsc{Random} & \multirow{4}{*}{5} & 0.09 & 0.27 & 0.27 & 0.27 & 0.27 & 0.27 & 0.27 \\
        \lr{} (ours) && 0.00 & 6.37 & 1.11 & 7.27 & 1.91 & 2.92 & 0.69 \\
        \lrc{} (ours) && \textbf{2.69} & \textbf{23.67} & \textbf{17.94} & \textbf{20.70} & \textbf{5.73} & \textbf{19.59} & \textbf{8.49} \\
        \lc{} (ours) && 1.65 & 13.22 & 8.39 & 10.08 & 3.18 & 10.46 & 5.47 \\
        \cmidrule(lr){1-9}
        \textsc{Random} & \multirow{4}{*}{10} & 0.18 & 0.53 & 0.53 & 0.53 & 0.53 & 0.53 & 0.53 \\
        \lr{} (ours) && 0.00 & 9.66 & 2.39 & 10.99 & 3.24 & 4.88 & 1.80 \\
        \lrc{} (ours) && \textbf{4.87} & \textbf{33.65} & \textbf{26.38} & \textbf{27.55} & \textbf{9.50} & \textbf{28.13} & \textbf{12.74} \\
        \lc{} (ours) && 3.01 & 19.53 & 13.48 & 15.29 & 5.57 & 15.45 & 8.39 \\
        \bottomrule
    \end{tabular}}
\end{table}

\section{Implementation and Hardware}
\label{app:implementation}

Our framework implementation is built upon the codebases by \citet{ramirez2024deep} (for the decoder and training) and \citet{lim2024graph} (for the GMN architecture and graph conversion). \nerf{}s belonging to \hash{}, \triob{}, \hashob{}, and the unseen architectures were trained with the NerfAcc framework \citep{li2022nerfacc}, as done in \citet{ramirez2024deep} and \citet{cardace2024neural} for \mlp{} and \tri{}. All our experiments were performed on a single NVIDIA RTX A6000. Training either \lr{} or \lrc{} took ${\sim}2$ weeks, whereas training \lc{} took ${\sim}4$ days.

\section{Additional Qualitative Results}
\label{app:qualitatives}

\begin{figure}
    \centering
    \includegraphics[width=\linewidth]{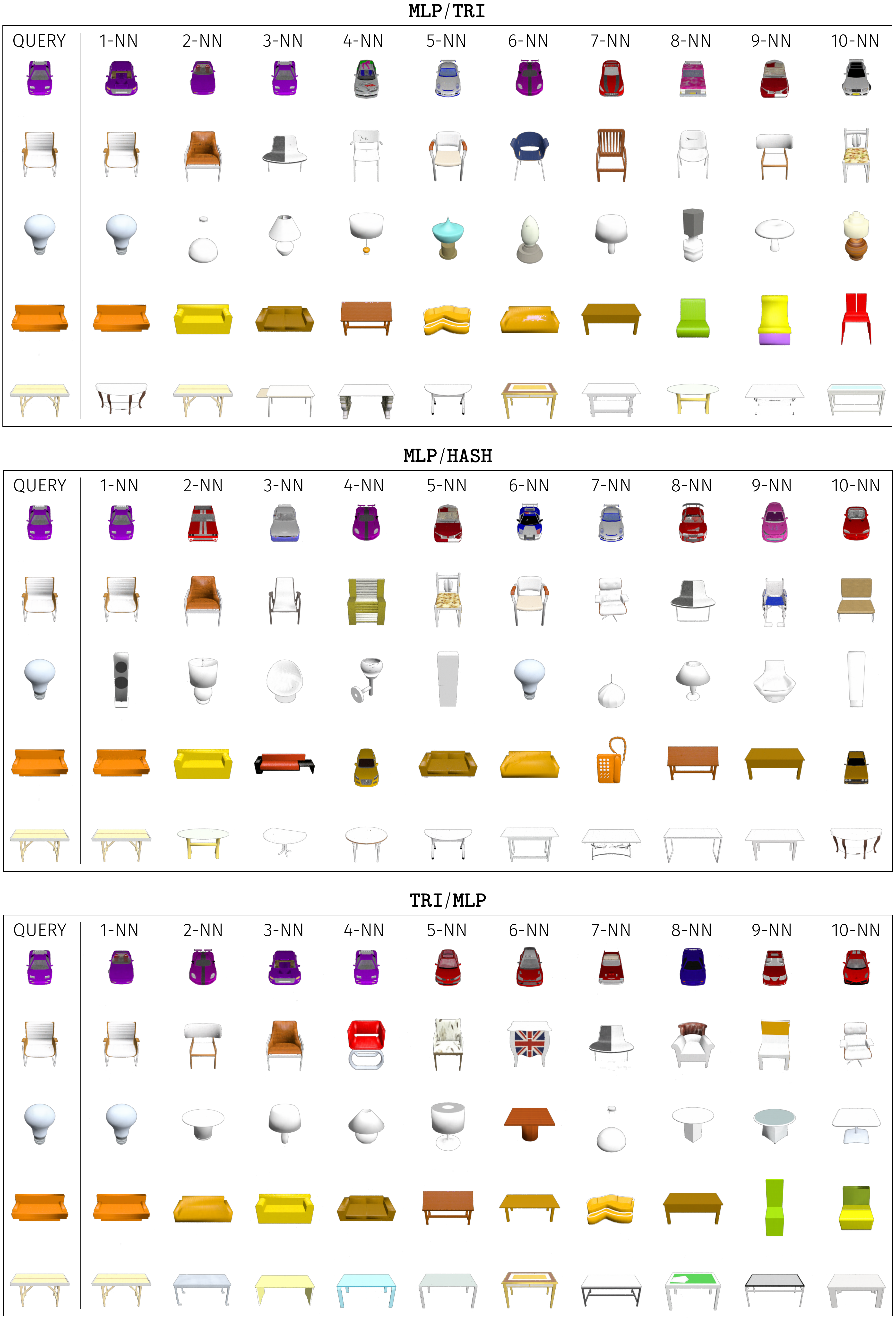}
    \caption{\textbf{\nerf{} retrieval qualitative results \normalfont{(\lrc{}).}} Query and gallery from the test set of \mlp{}, \tri{}, or \hash{}.}
    \label{fig:retrieval-full-1}
\end{figure}

\begin{figure}
    \centering
    \includegraphics[width=\linewidth]{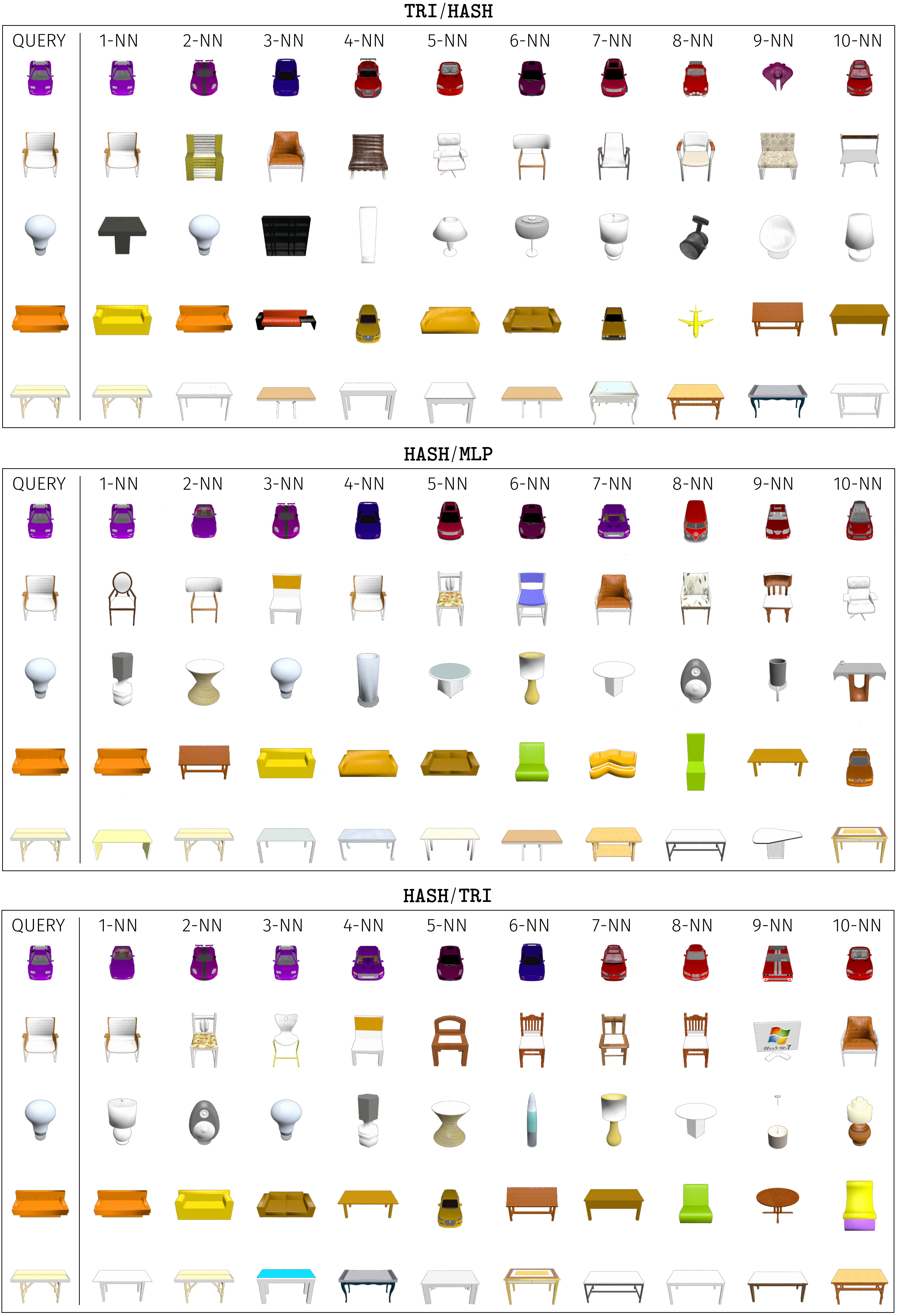}
    \caption{\textbf{\nerf{} retrieval qualitative results \normalfont{(\lrc{}).}} Query and gallery from the test set of \mlp{}, \tri{}, or \hash{}.}
    \label{fig:retrieval-full-2}
\end{figure}

\cref{fig:retrieval-full-1,fig:retrieval-full-2} show \nerf{} retrieval qualitative results produced by \lrc{} for varying query/gallery combinations belonging to \mlp{}, \tri{}, or \hash{}. The similarity in color and/or appearance between the first 10 nearest neighbors and the query suggests that our framework encodes both relevant information about the appearance of the object represented by the \nerf{} and some invariance to the architecture used to parameterize it.

\begin{figure}
    \centering
    \includegraphics[width=\linewidth]{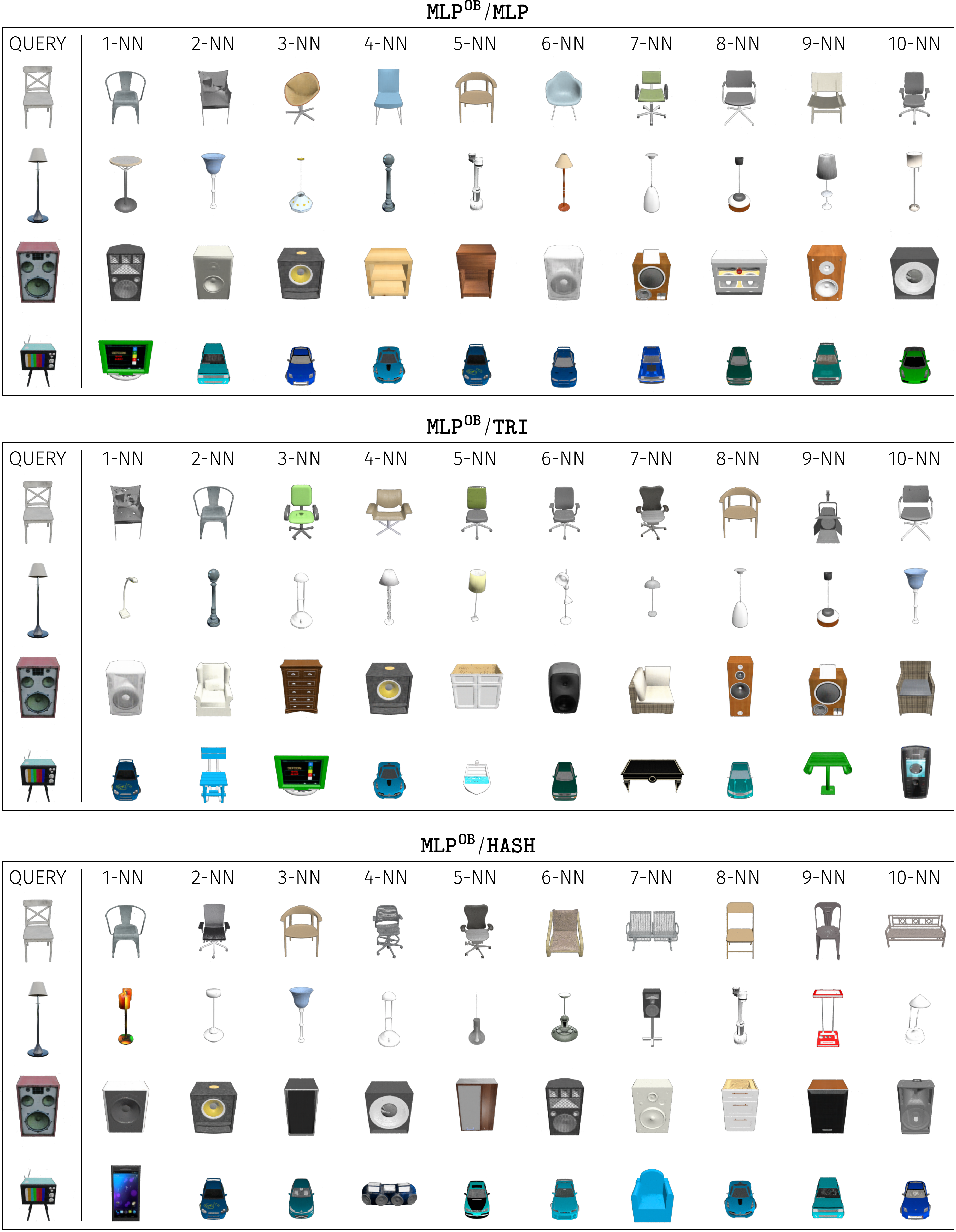}
    \caption{\textbf{Across-datasets \nerf{} retrieval \normalfont{(\lrc{}).}} Extended version of \cref{fig:retrieval}. Query from the test set of \mlpob{} \citep{amaduzzi2025scaling}, gallery from the test set of \mlp{}, \tri{}, or \hash{}.}
    \label{fig:retrieval-objaverse-full}
\end{figure}

\cref{fig:retrieval-objaverse-full} shows \nerf{} retrieval qualitative results produced by \lrc{} when the query belongs to \mlpob{} \citep{amaduzzi2025scaling} and the gallery belongs to \mlp{}, \tri{}, or \hash{}. As a reminder, \mlpob{}’s \nerf{}s are vanilla MLPs with the same architecture as those in \mlp{} (see \cref{tab:arch-app}). Since our framework was trained on \shapenet{} \citep{xu2019disn} NeRFs only, the purpose of this experiment is to shed some light on its ability to generalize to unseen datasets. The 10 nearest neighbors of the \mlpob{} query reveal how remarkably well our method can retrieve NeRFs belonging to the same class as the query when the latter belongs to \shapenet{} classes (i.e. \emph{chair}, \emph{lamp}, and \emph{speaker}), while mostly failing for the television query (whose closest \shapenet{} class would be \emph{display}), although, interestingly, at least one object featuring a display with a pattern similar to that of the query is always found among the closest neighbors.

\begin{figure}
    \centering
    \includegraphics[width=\linewidth]{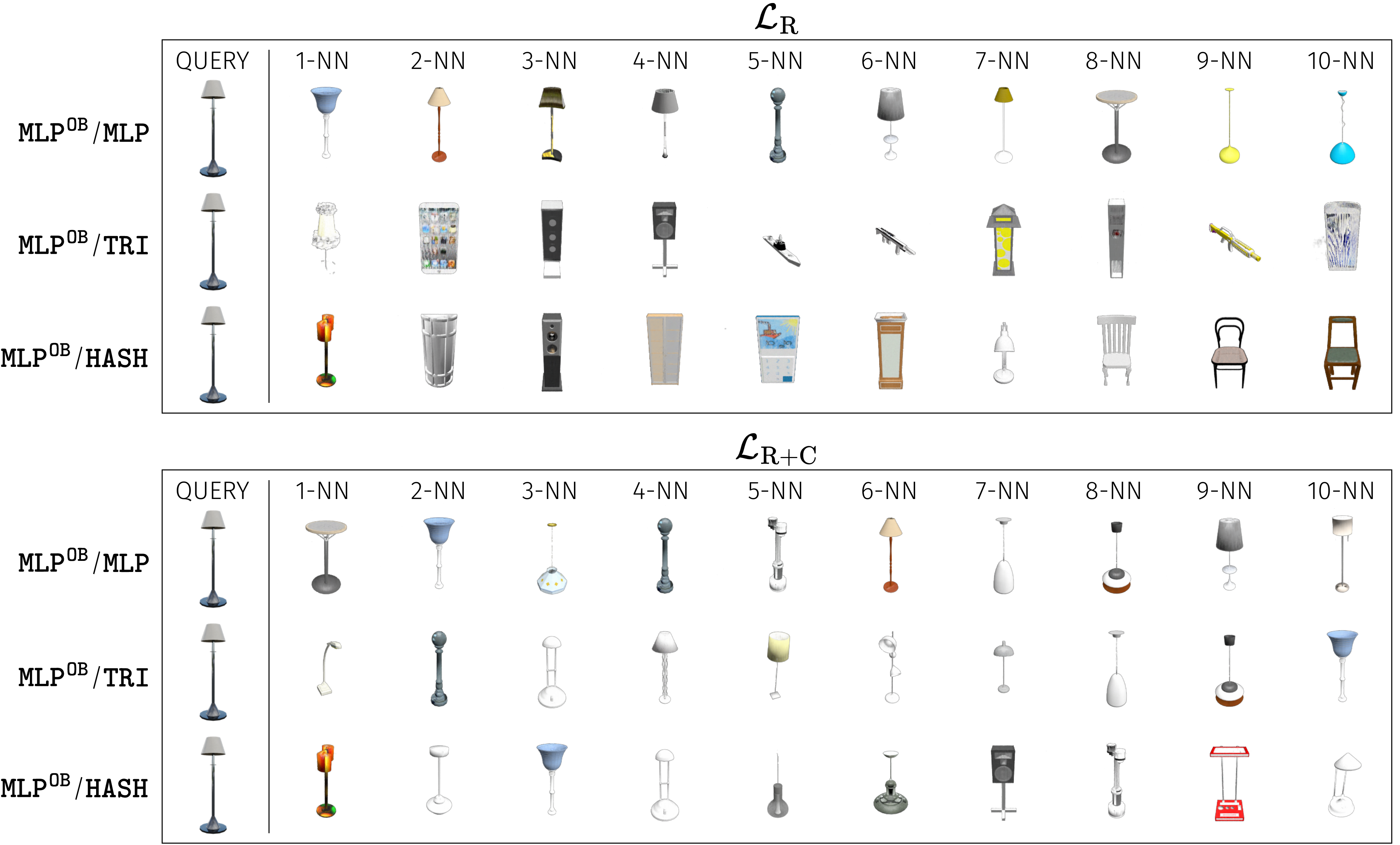}
    \caption{\textbf{Across-datasets \nerf{} retrieval \normalfont{(\lr{} vs \lrc{}).}} Query from the test set of \mlpob{} \citep{amaduzzi2025scaling}, gallery from the test set of \mlp{}, \tri{}, or \hash{}.}
    \label{fig:retrieval-objaverse-lamp}
\end{figure}

\cref{fig:retrieval-objaverse-lamp} compares retrieval qualitative results of \lr{} and \lrc{} for one \mlpob{} \citep{amaduzzi2025scaling} query. As one may expect, the contrastive term in \lrc{} allows retrieving objects belonging to the same class as the query even when the gallery \nerf{}s are parameterized by different architectures than the query (i.e.\ \tri{} and \hash{}), whereas \lr{} produces reasonable results for all 10 nearest neighbors only when the gallery architecture is the same as the query (i.e.\ \mlp{}), which, once again, demonstrates the importance of the contrastive objective when dealing with multiple architectures.

\section{Additional \nerf{} Captioning and Q\&A Results}
\label{app:language}

\cref{tab:caption-brief-multi-full,tab:caption-detailed-multi-full,tab:q&a-multi-full,tab:caption-brief-single-full,tab:caption-detailed-single-full,tab:q&a-single-full} show extended results for the language tasks described in \cref{sec:language} by adding BLEU-1 \citep{papineni2002bleu}, ROUGE-L \citep{chin2004rouge}, and METEOR \citep{banerjee2005meteor} metrics, which are traditional handcrafted similarity measures based on n-gram statistics. As discussed in \cref{sec:language}, S-BERT \citep{reimers2019sentence} and SimCSE \citep{gao2021simcse} are instead based on pre-trained networks, and generally considered more effective than n-gram-based metrics at measuring the quality of the generated text \citep{amaduzzi2024llana}.

\begin{table}[t]
    \centering
    \begin{minipage}[t]{0.9\linewidth}
        \caption{\textbf{\nerf{} brief captioning (multi-architecture)}. Extended version of \cref{tab:caption-brief-multi}.}
        \label{tab:caption-brief-multi-full}
        \centering
        \resizebox{\linewidth}{!}{
        \begin{tabular}{lcccccc}
            \toprule
            Method & Test Dataset & S-BERT $\uparrow$ & SimCSE $\uparrow$ & BLEU-1 $\uparrow$ & ROUGE-L $\uparrow$ & METEOR $\uparrow$ \\
            \midrule
            \llana{} \citep{amaduzzi2024llana} & -- & -- & -- & -- & -- & -- \\
            \cmidrule(lr){1-7}
            \multirow{3}{*}{\lrc{} (ours)} & \mlp{} & \textbf{67.1} & \textbf{68.9} & \textbf{32.7} & \textbf{35.1} & \textbf{36.1} \\
            & \tri{} & 66.4 & 67.8 & 32.5 & 34.4 & 35.9 \\
            & \hash{} & 66.3 & 67.9 & 32.4 & 34.5 & 35.6 \\
            \bottomrule
        \end{tabular}}
        \end{minipage}
    \hfill
    \begin{minipage}[t]{0.9\linewidth}
        \caption{\textbf{\nerf{} detailed captioning (multi-architecture)}. Extended version of \cref{tab:caption-detailed-multi}.}
        \label{tab:caption-detailed-multi-full}
        \centering
        \resizebox{\linewidth}{!}{
        \begin{tabular}{lcccccc}
            \toprule
            Method & Test Dataset & S-BERT $\uparrow$ & SimCSE $\uparrow$ & BLEU-1 $\uparrow$ & ROUGE-L $\uparrow$ & METEOR $\uparrow$ \\
            \midrule
            \llana{} \citep{amaduzzi2024llana} & -- & -- & -- & -- & -- & -- \\
            \cmidrule(lr){1-7}
            \multirow{3}{*}{\lrc{} (ours)} & \mlp{} & \textbf{73.3} & \textbf{75.1} & \textbf{19.7} & \textbf{32.1} & \textbf{20.1} \\
            & \tri{} & 72.2 & 73.8 & 19.5 & 31.4 & 19.7 \\
            & \hash{} & 72.4 & 74.0 & 19.6 & 31.7 & 19.9 \\
            \bottomrule
        \end{tabular}}
    \end{minipage}
    \hfill
    \begin{minipage}[t]{0.9\linewidth}
        \caption{\textbf{\nerf{} single-round Q\&A (multi-architecture)}. Extended version of \cref{tab:q&a-multi}.}
        \label{tab:q&a-multi-full}
        \centering
        \resizebox{\linewidth}{!}{
        \begin{tabular}{lcccccc}
            \toprule
            Method & Test Dataset & S-BERT $\uparrow$ & SimCSE $\uparrow$ & BLEU-1 $\uparrow$ & ROUGE-L $\uparrow$ & METEOR $\uparrow$ \\
            \midrule
            \llana{} \citep{amaduzzi2024llana} & -- & -- & -- & -- & -- & -- \\
            \cmidrule(lr){1-7}
            \multirow{3}{*}{\lrc{} (ours)} & \mlp{} & \textbf{80.9} & \textbf{81.4} & 45.6 & 52.6 & 49.5 \\
            & \tri{} & 80.7 & 81.3 & \textbf{45.7} & \textbf{52.9} & \textbf{49.8} \\
            & \hash{} & 80.6 & 81.3 & 45.3 & 52.4 & 49.3 \\
            \bottomrule
        \end{tabular}}
    \end{minipage}
    \hfill

    \begin{minipage}[t]{0.9\linewidth}
        \caption{\textbf{\nerf{} brief captioning (single-architecture)}. Extended version of \cref{tab:caption-brief-single}.}
        \label{tab:caption-brief-single-full}
        \centering
        \resizebox{\linewidth}{!}{
        \begin{tabular}{lccccc}
            \toprule
            Method & S-BERT $\uparrow$ & SimCSE $\uparrow$ & BLEU-1 $\uparrow$ & ROUGE-L $\uparrow$ & METEOR $\uparrow$ \\
            \midrule
            \llana{} \citep{amaduzzi2024llana} & \textbf{68.6} & \textbf{70.5} & 20.6 & 28.3 & 31.8 \\
            \lr{} (ours) & 68.3 & 70.2 & \textbf{33.1} & \textbf{35.3} & \textbf{36.6} \\
            \bottomrule
        \end{tabular}}
        \end{minipage}
    \hfill
    \begin{minipage}[t]{0.9\linewidth}
        \caption{\textbf{\nerf{} detailed captioning (single-architecture)}. Extended version of \cref{tab:caption-detailed-single}.}
        \label{tab:caption-detailed-single-full}
        \centering
        \resizebox{\linewidth}{!}{
        \begin{tabular}{lccccc}
            \toprule
            Method & S-BERT $\uparrow$ & SimCSE $\uparrow$ & BLEU-1 $\uparrow$ & ROUGE-L $\uparrow$ & METEOR $\uparrow$ \\
            \midrule
            \llana{} \citep{amaduzzi2024llana} & \textbf{77.4} & \textbf{79.8} & \textbf{41.3} & \textbf{36.2} & \textbf{32.4} \\
            \lr{} (ours) & 74.1 & 76.0 & 20.1 & 32.5 & 20.6 \\
            \bottomrule
        \end{tabular}}
    \end{minipage}
    \hfill
    \begin{minipage}[t]{0.9\linewidth}
        \caption{\textbf{\nerf{} single-round Q\&A (single-architecture)}. Extended version of \cref{tab:q&a-single}.}
        \label{tab:q&a-single-full}
        \centering
        \resizebox{\linewidth}{!}{
        \begin{tabular}{lccccc}
            \toprule
            Method & S-BERT $\uparrow$ & SimCSE $\uparrow$ & BLEU-1 $\uparrow$ & ROUGE-L $\uparrow$ & METEOR $\uparrow$ \\
            \midrule
            \llana{} \citep{amaduzzi2024llana} & \textbf{81.0} & \textbf{81.6} & \textbf{46.2} & \textbf{53.2} & \textbf{50.2} \\
            \lr{} (ours) & \textbf{81.0} & \textbf{81.6} & 45.4 & 52.5 & 49.4 \\
            \bottomrule
        \end{tabular}}
    \end{minipage}
\end{table}

\section{LLM Usage}
LLMs were used in this paper as a tool to aid and polish writing. No significant role in research ideation, experimentation, or writing was delegated to LLMs.

\end{document}